\pdfoutput=1
\def\year{2017}\relax
\documentclass[letterpaper]{article}
\usepackage{aaai17}
\usepackage[table,x11names]{xcolor}
\usepackage{times}
\usepackage{helvet}
\usepackage{courier}
\usepackage[utf8]{inputenc} %
\usepackage[T1]{fontenc}    %
\usepackage{url}            %
\usepackage{booktabs}       %
\usepackage{amsfonts}       %
\usepackage{nicefrac}       %
\usepackage{microtype}      %
\usepackage{times}
\usepackage{graphicx}
\usepackage{sidecap}
\usepackage{algorithm}
\usepackage{algorithmic}
\usepackage{subcaption}
\usepackage{cancel}
\usepackage{amsmath}
\usepackage{dsfont}
\usepackage{amsthm}
\usepackage{tikz-qtree}
\usepackage{listings}
\usepackage{bbm}
\usepackage{mathtools}
\usepackage{breqn}
\usepackage{todonotes}
\usepackage{rotating}
\usepackage{mathtools}
\usepackage{ amssymb }
\usepackage{multirow}

\newcommand{\mathtext}[1]{\small{\textit{(#1)}}}
\newcommand{\Exp}[2]{\mathop{\mathbb{E}}_{#2}\left[#1\right]}
\newcommand{\vsigma}{\vec{\sigma}}
\DeclareMathOperator{\diag}{diag}
\DeclareMathOperator{\KL}{KL}

\newcommand{\pth}{p_\theta}
\newcommand{\qph}{q_\phi}
\newcommand{\Id}{\mathds{1}}
\newcommand{\dt}{\Delta}

\DeclareMathOperator{\Tr}{Tr}
\newcommand{\Prec}[1]{{\Sigma}_{#1}^{-1}}

\newcommand{\vecx}{\vec{x}}

\newcommand{\vecz}{\vec{z}}
\newcommand{\bigCI}{\mathrel{\text{\scalebox{1.07}{$\perp\mkern-10mu\perp$}}}}

\newcommand{\meanfxn}{\textit{G}_{\alpha}}
\newcommand{\covfxn}{\textit{S}_{\beta}}
\newcommand{\emisfxn}{\textit{F}_{\kappa}}
\newcommand{\lthph}{\mathcal{L}(\vecx;(\theta, \phi))}

\usepackage{tikz}
\usetikzlibrary{bayesnet}
\usetikzlibrary{calc}
\theoremstyle{plain}

\newcommand{\MLP}{\text{MLP}}
\newcommand{\ReLU}{\text{ReLU}}
\newcommand{\Tanh}{\text{tanh}}
\newcommand{\Sigmoid}{\text{sigmoid}}
\newcommand{\Softplus}{\text{softplus}}
\newcommand{\citep}{\cite}
\newcommand{\citet}{\cite}
\usepackage{xspace}
\newcommand{\DKF}{DMM\xspace}
\newcommand{\DMM}{DMM\xspace}
\newcommand{\DKS}{\textbf{DKS}\xspace}

\title{Title}
\author{Rahul G. Krishnan, Uri Shalit, David Sontag\\
	Courant Institute of Mathematical Sciences, 
	New York University\\
	\{rahul, shalit, dsontag\}@cs.nyu.edu\\
}
\begin{document}
\title{Structured Inference Networks for Nonlinear State Space Models} 
\maketitle
\begin{abstract}
Gaussian state space models have been used for decades as generative models of sequential data. They admit an intuitive probabilistic interpretation, have a simple functional form, and enjoy widespread adoption. We introduce a unified algorithm to efficiently learn a broad class of linear and non-linear state space models, including variants where the emission and transition distributions are modeled by deep neural networks. Our learning algorithm simultaneously learns a compiled inference network and the generative model, leveraging a structured variational approximation parameterized by recurrent neural networks to mimic the posterior distribution. We apply the learning algorithm to both synthetic and real-world datasets, demonstrating its scalability and versatility. We find that using the structured approximation to the posterior results in models with significantly higher held-out likelihood.

\end{abstract}

\section{Introduction}

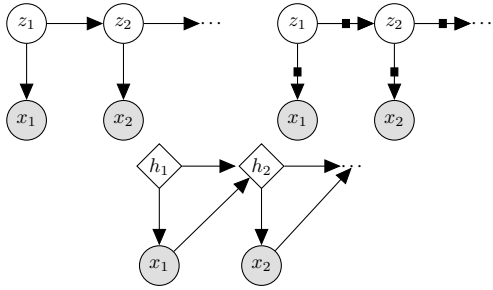
\begin{figure}[t]
\centering
	\centering
	\begin{tikzpicture}[scale=0.75, transform shape,blackdot/.style={thin, draw=black, align=center, scale = 0.3,fill=black}]
	\node [latent] (z1) {$z_1$};
	\node [latent, right= of z1] (z2) {$z_2$};
	\node [const, right=of z2] (dotsz) {\ldots};
	\node [obs, below= of z1] (x1) {$x_1$};
	\node [obs, below= of z2] (x2) {$x_2$};
	\edge {z1} {z2};
	\edge {z2} {dotsz};
	\edge {z1} {x1};
	\edge {z2} {x2};
	\end{tikzpicture}\qquad
	\begin{tikzpicture}[scale=0.75, transform shape,blackdot/.style={thin, draw=black, align=center, scale = 0.3,fill=black}]
	\node [latent] (z1) {$z_1$};
	\node [latent, right= of z1] (z2) {$z_2$};
	\node [const, right=of z2] (dotsz) {\ldots};
	\node [obs, below= of z1] (x1) {$x_1$};
	\node [obs, below= of z2] (x2) {$x_2$};
	\edge {z1} {z2};
	\edge {z2} {dotsz};
	\draw[->]  (z1) -- node[blackdot] {d} (z2);
	\draw[->]  (z1) -- node[blackdot] {d} (x1);
	\draw[->]  (z2) -- node[blackdot] {d} (x2);
	\draw[->]  (z2) -- node[blackdot] {d} (dotsz);
	\end{tikzpicture}\quad
	\begin{tikzpicture}[scale=0.75, transform shape,blackdot/.style={thin, draw=black, align=center, scale = 0.3, fill=black}]
	\node [det] (h1) {$h_1$};
	\node [det, right= of h1] (h2) {$h_2$};
	\node [const, right=of h2] (dotsz) {\ldots};
	\node [obs, below= of h1] (x1) {$x_1$};
	\node [obs, below= of h2] (x2) {$x_2$};
	\edge {h1} {z2};
	\edge {h2} {dotsz};
	\edge {x2} {dotsz};
	\edge {h1} {x1};
	\edge {x1} {h2};
	\edge {h2} {x2};
	\end{tikzpicture}
	\caption{\small \textbf{Generative Models of Sequential Data: } (\textbf{Top Left}) Hidden Markov Model (HMM), (\textbf{Top Right}) Deep Markov Model (DMM)
	{ {$\blacksquare$}} denotes the neural networks used in DMMs
        for the emission and transition functions. 
	(\textbf{Bottom}) Recurrent Neural Network (RNN), {{$\Diamond$}} denotes a deterministic intermediate representation. 
	Code for learning
        DMMs and reproducing our results 
	may be found at: \texttt{\small github.com/clinicalml/structuredinference} \label{fig:dkf}}
\end{figure}

Models of sequence data such as hidden Markov models (HMMs) and
recurrent neural networks (RNNs)
are widely used in machine translation, 
speech recognition, 
and computational biology.
Linear and non-linear
Gaussian state space models (GSSMs, Fig.~\ref{fig:dkf}) are used in applications including
robotic planning and missile
tracking. 
However, despite huge progress over the last decade, efficient learning of
non-linear models from
complex high dimensional time-series remains a major challenge.
Our paper proposes a unified learning algorithm for a broad class of
GSSMs, and we introduce an inference procedure that scales easily to high dimensional data, 
compiling approximate (and where feasible, exact)
inference into the parameters of a neural network.  

In engineering and control, the parametric form of the GSSM model is often known, with typically a few
specific parameters that need to be fit to data. The most commonly used approaches for these types of learning and inference problems are often computationally demanding, e.g. dual extended Kalman filter \cite{Wan_NIPS96}, expectation maximization \cite{Briegel99fisherscoring,roweis2000algorithm}
or particle filters \cite{schon2011system}. Our compiled inference algorithm can easily deal with high-dimensions both in the observed and the latent spaces, without compromising the quality of inference and learning.

When the parametric form of the model is unknown, 
we propose learning {\em deep Markov models} (\DKF), a class of generative models 
where classic linear emission and
transition distributions 
are replaced with complex multi-layer perceptrons (MLPs).
These are GSSMs that retain the 
Markovian structure of HMMs, but
leverage the representational power of deep neural 
networks to model complex high dimensional data. 
If one augments a \DKF model such as the one presented in Fig.~\ref{fig:dkf} with edges from the observations $x_t$
to the latent states of the following time step $z_{t+1}$, then the \DKF 
 can be seen to be similar to, though more restrictive than, stochastic RNNs \cite{bayer2014learning} and variational RNNs
\cite{chung2015recurrent}.

Our learning algorithm performs stochastic gradient ascent on
a variational lower bound of the likelihood. 
Instead of introducing
variational parameters for each data point, we {\em compile} the
inference procedure at the same time as learning the generative model. 
This idea was originally used in the wake-sleep algorithm for
unsupervised learning \cite{hinton1995wake}, and has since led to 
state-of-the-art results for unsupervised learning of deep
generative models \cite{kingma2013auto,mnih2014neural,rezende2014stochastic}. 

Specifically, we introduce a new family of {\em structured
 inference networks}, parameterized by recurrent neural networks, and
evaluate their effectiveness in three scenarios: 
(1) when the generative model is known and fixed, 
(2) in parameter estimation when the functional form of the model is known 
and (3) for learning deep Markov models.
By looking at the structure of the true posterior, we show both
theoretically and empirically that inference for a latent state should
be performed using information \emph{from its future}, as opposed to
recent work which performed inference using only information from the
past \cite{chung2015recurrent,gan2015deep,gregor2015draw}, and that a
structured variational approximation outperforms mean-field based approximations.
Our approach may easily be 
adapted to learning more general generative models, for example models
with edges from observations to latent states.  

Finally, we learn a \DMM on a polyphonic music dataset and on a dataset of
electronic health records (a complex high dimensional setting with missing data). We use the model learned on health records to ask queries such as ``what would have happened to patients had they not received treatment'', and show that our model correctly identifies the way certain medications affect a patient's health.

\textbf{Related Work:} Learning GSSMs with MLPs for the transition distribution was considered by \cite{raiko2009variational}. 
They approximate the posterior with non-linear dynamic 
factor analysis \cite{valpola2002unsupervised}, which scales quadratically with 
the observed dimension and is impractical for large-scale learning.

Recent work has considered variational learning of time-series
data using structured inference or recognition networks.
\citeauthor{archer2015black} propose using a
Gaussian approximation to the posterior distribution with a block-tridiagonal inverse covariance. 
\citeauthor{johnson2016structured} use a conditional random field as the inference network for time-series models. 
Concurrent to our own work, \citeauthor{fraccaro2016sequential} also learn sequential generative models 
using structured inference networks parameterized by recurrent neural networks.

\citeauthor{bayer2014learning} and \citeauthor{fabius2014variational} create a stochastic variant of RNNs 
by making the hidden state of the RNN at every time step be a function
of independently sampled latent variables. \citeauthor{chung2015recurrent}
apply a similar model to speech data, sharing parameters between the
RNNs for the generative model and the inference network.
\citeauthor{gan2015deep} learn a 
model with discrete random variables, using a structured inference network that only considers information from the 
past, similar to \citeauthor{chung2015recurrent} and \citeauthor{gregor2015draw}'s models. 
In contrast to these works, we use information from the future within a
structured inference network, which we show to be preferable both
theoretically and practically. Additionally, we systematically
evaluate the impact of the different variational approximations on
learning. %

\citeauthor{watter2015embed} construct a first-order Markov model using inference networks. However, their learning
algorithm is based on data tuples over consecutive time steps. This makes the strong assumption
that the posterior distribution can be recovered based on observations
at the current and next time-step.
As we show, for generative models like the one in Fig. \ref{fig:dkf}, the 
posterior distribution at any time step is a function of \emph{all}
future (and past) observations. 

\section{Background}\label{sec:back}

\textbf{Gaussian State Space Models:}
We consider both inference and learning in a class of latent variable models given by:
We denote by $z_t$ a vector valued latent variable and by $x_t$ a vector valued observation. 
A sequence of such latent variables and observations is denoted $\vec{z},\vec{x}$ respectively.
\begin{align}
	\label{eqn:gen_model}
	&z_t \sim \mathcal{N} (\meanfxn(z_{t-1},\dt_t),\covfxn(z_{t-1},\dt_t)) &\mathtext{Transition}\\ 
	&x_t \sim \Pi(\emisfxn(z_t))\; &\mathtext{Emission} \;\;
\end{align}
We assume that the distribution of the latent states is a multivariate
Gaussian with a mean and covariance which are differentiable functions of the previous latent state and $\dt_t$ (the time elapsed
of time between $t-1$ and $t$). 
The multivariate observations $x_t$ are distributed according to a distribution $\Pi$ 
(e.g., independent Bernoullis if the data is binary) whose parameters are a function of the corresponding latent state $z_t$.
Collectively, we denote by $\theta=\{\alpha,\beta,\kappa\}$ the parameters of the generative model. 

Eq. \ref{eqn:gen_model} subsumes a large family of linear 
and non-linear Gaussian state space models. 
For example, by setting $\meanfxn(z_{t-1}) = G_t z_{t-1},\covfxn = \Sigma_t,\emisfxn=F_t z_{t}$, where $G_t$, $\Sigma_t$ and $F_t$ are matrices,
we obtain linear state space models. 
The functional forms
and initial parameters for $\meanfxn,\covfxn,\emisfxn$ may be pre-specified.

\textbf{Variational Learning: }
Using recent advances in variational inference 
we optimize a variational lower bound on the data log-likelihood. 
The key technical innovation is the introduction of an \emph{inference network} or \emph{recognition network} \cite{hinton1995wake,kingma2013auto,mnih2014neural,rezende2014stochastic}, a neural network which approximates the intractable posterior. This is a parametric conditional distribution that is optimized to perform inference. 
Throughout this paper we will use $\theta$ to denote the parameters of the generative model, and $\phi$ to denote the parameters of the inference network.

For the remainder of this section, we consider learning in a Bayesian network whose joint distribution factorizes as:
$p(x,z) = \pth(z) \pth(x|z)$. 
The posterior distribution $\pth(z|x)$ is typically intractable. Using the well-known 
variational principle, we posit an approximate posterior distribution $\qph(z|x)$ to
obtain the following lower bound on the marginal likelihood:
\begin{equation}\label{eqn:varlowbnd}
 \log \pth(x) \geq \Exp{\log \pth(x|z)}{\qph(z|x)} - \KL(\, \qph(z|x) || \pth(z)\, ),
\end{equation}
where the inequality is by Jensen's inequality. 
\citeauthor{kingma2013auto,rezende2014stochastic} use a neural net (with parameters $\phi$) to parameterize $\qph$.
The challenge in the resulting optimization problem is that the lower bound in Eq. \ref{eqn:varlowbnd} includes an expectation w.r.t. $\qph$, which implicitly depends on the network parameters $\phi$. 
When using a Gaussian variational approximation 
$\qph(z|x) \sim \mathcal{N}\left(\mu_\phi(x),\Sigma_\phi(x)\right)$, 
where $\mu_\phi(x),\Sigma_\phi(x)$ are parametric functions of the
observation $x$, this difficulty is overcome by using \emph{stochastic backpropagation}:
a simple transformation allows 
one to obtain unbiased Monte Carlo estimates of the gradients of $\Exp{\log \pth(x|z)}{\qph(z|x)} $ with respect to $\phi$.
The $\KL$ term in Eq. \ref{eqn:varlowbnd} can be estimated similarly since it is also an expectation. 
When the prior $\pth(z)$ is Normally distributed, the $\KL$ and its gradients 
may be obtained analytically.

\section{A Factorized Variational Lower Bound \label{sec:learnmodel}}

We leverage stochastic backpropagation to learn generative models given by 
Eq.~\ref{eqn:gen_model}, corresponding to the graphical model in Fig. \ref{fig:dkf}.
Our insight is that for the purpose of inference, we can use the Markov properties of
the generative model to guide us in deriving a structured
approximation to the posterior. Specifically, the posterior factorizes as:
\begin{equation}
	\label{thm:p_fact}
	p(\vecz|\vecx) = p(z_1|\vecx)\prod_{t=2}^T p(z_t|z_{t-1},x_t,\ldots,x_T).
\end{equation}
To see this, use the independence statements implied by the graphical
model in Fig. \ref{fig:dkf}
 to note that $p(\vecz|\vecx)$, the true posterior, factorizes as:
	\begin{equation*} p(\vecz|\vecx) =
          p(z_1|\vecx)\prod_{t=2}^T
          p(z_t|z_{t-1},\vecx) 
\end{equation*}
Now, we notice that $z_t\bigCI x_1,\ldots,x_{t-1}|z_{t-1}$, yielding the desired result.
The significance of Eq. \ref{thm:p_fact} is that it yields 
insight into the structure of the exact posterior for the class of models
laid out in Fig. \ref{fig:dkf}. 

We directly mimic the structure of the posterior with the following factorization of the variational approximation:
\begin{align}
	\label{eqn:q_fact}
&\qph(\vecz|\vecx) = \qph(z_1|x_1,\ldots,x_T) \prod_{t=2}^T \qph(z_t|z_{t-1},x_t, \ldots, x_T)\\ 
&\text{s.t.}\quad \qph(z_t|z_{t-1},x_t, \ldots, x_T) \sim \nonumber\\
& \quad \quad \quad \mathcal{N}\left(\mu_\phi(z_{t-1},x_t, \ldots, x_T), \Sigma_\phi(z_{t-1},x_t, \ldots, x_T)\right) \nonumber
\end{align}
where $\mu_\phi$ and $\Sigma_\phi$ are functions parameterized by neural nets.
Although $\qph$ has the option to condition on all information across time, Eq. \ref{thm:p_fact}
suggests that in fact it suffices to condition on information from the future
and the previous latent state. The previous latent state serves as a summary statistic for information from the past.

\textit{Exact Inference: } We can match the factorization of the true posterior using the inference network
but using a Gaussian variational approximation for the approximate posterior over each latent variable (as we do) limits the expressivity
of the inferential model, except for the case of linear dynamical
systems where the posterior distribution is Normally distributed.
However, one could augment our proposed inference network with recent innovations that improve the variational
approximation to allow for multi-modality \cite{rezende2015variational,tran2016variational}. Such modifications could yield black-box 
methods for exact inference in time-series models, which we leave for future work.

{\bf Deriving a Variational Lower Bound:} 
For a generative model (with parameters $\theta$) and an inference network (with parameters $\phi$), we are interested in 
$\max_{\theta} \log \pth(\vecx)$. For ease of exposition, we instantiate the derivation of the variational bound for a single data point $\vec{x}$
though we learn $\theta,\phi$ from a corpus. 

The lower bound in Eq.~\ref{eqn:varlowbnd} has an analytic form of the $\KL$ term only for the simplest of transition models $\meanfxn,\covfxn$ between $z_{t-1}$ and $z_t$ (Eq. \ref{eqn:gen_model}). 
One could estimate the gradient of the KL term by sampling from the variational model, 
but that results in high variance estimates and gradients. We use a different factorization of the KL term (obtained by using the prior distribution over latent variables), leading to the variational lower bound we use as our objective function:
\begin{align}
\label{eqn:bound_likelihood}
&\lthph = \sum_{t=1}^T\Exp{\log \pth(x_t|z_t)}{\qph(z_t|\vecx)}\\ 
&-\KL(\qph(z_1|\vecx)||\pth(z_1))\nonumber\\
&-\sum_{t=2}^{T} \Exp{\KL(\qph(z_t|z_{t-1},\vecx)||\pth(z_t|z_{t-1}))}{\qph(z_{t-1}|\vecx)}\nonumber.
\end{align}

The key point is the resulting objective function has more stable analytic gradients.
Without the factorization of the KL divergence in Eq. \ref{eqn:bound_likelihood}, 
we would have to estimate $\KL(q(\vec{z}|\vec{x})||p(\vec{z}))$ via
Monte-Carlo sampling, since it has no analytic form. In contrast, in Eq. \ref{eqn:bound_likelihood} 
the individual KL terms \emph{do} have analytic forms.
A detailed derivation of the bound and the factorization of the KL divergence is detailed in the supplemental material.

\textbf{Learning with Gradient Descent:} 
The objective in Eq.~\ref{eqn:bound_likelihood} is differentiable
in the parameters of the model ($\theta,\phi$).
If the generative model $\theta$ is fixed, we perform gradient ascent of Eq. \ref{eqn:bound_likelihood}
in $\phi$. Otherwise, we perform gradient ascent in both $\phi$ and $\theta$.
We use stochastic backpropagation \cite{kingma2013auto,rezende2014stochastic} 
for estimating the gradient w.r.t. $\phi$. 
Note that the expectations are only taken with respect to the variables $z_{t-1}, z_t$, which are the sufficient statistics of the Markov model. 
For the KL terms in Eq. \ref{eqn:bound_likelihood}, we use the fact that the prior $\pth(z_t|z_{t-1})$ and the
variational approximation to the posterior
$\qph(z_t|z_{t-1},\vecx)$ are both Normally distributed, and
hence their KL divergence may be estimated analytically.

\begin{algorithm}[h]
\caption{\small \textbf{Learning a \DMM with stochastic gradient descent: } 
	We use a single sample from the recognition network during learning to evaluate expectations in the bound. 
We aggregate gradients across mini-batches.}
\begin{algorithmic} \label{alg1}
	\STATE \textbf{Inputs}: Dataset $\mathcal{D}$
\STATE \qquad\quad\; Inference Model: $\qph(\vecz|\vecx)$
\STATE \qquad\quad\; Generative Model: $\pth(\vecx|\vecz), \pth(\vecz)$
\WHILE{$notConverged()$}
\STATE 1. Sample datapoint: $\vecx\sim\mathcal{D}$
\STATE 2. Estimate posterior parameters (Evaluate $\mu_\phi, \Sigma_\phi$)
\STATE 3. Sample $\hat{\vecz}\sim \qph(\vecz|\vecx)$
\STATE 4. Estimate conditional likelihood: $\pth(\vecx|\hat{\vecz})$ \& KL
\STATE 5. Evaluate $\lthph$ 
\STATE 6. Estimate MC approx. to $\nabla_{\theta}\mathcal{L}$ 
\STATE 7. Estimate MC approx. to $\nabla_{\phi}\mathcal{L}$ 
\STATE (Use stochastic backpropagation to move gradients with respect to $\qph$ inside expectation)
\STATE 8. Update $\theta,\phi$ using ADAM \citep{kingma2014adam} 
\ENDWHILE 
\end{algorithmic}
\end{algorithm}

Algorithm \ref{alg1} depicts an overview of the learning algorithm.
We outline the algorithm for a mini-batch of size one, 
but in practice gradients are averaged across stochastically sampled mini-batches of the training set.
We take a gradient step in $\theta$ and
$\phi$, typically with an adaptive learning rate such as
\citet{kingma2014adam}. 

\section{Structured Inference Networks\label{sec:opt_q}}

We now detail how 
we construct the variational approximation $q_\phi$, and specifically 
how we model the mean and diagonal covariance functions $\mu$ and $\Sigma$ using recurrent neural networks (RNNs).
Since our implementation only models the diagonal of the covariance matrix (the vector valued variances), we denote this as $\sigma^2$ rather
than $\Sigma$.
This parameterization cannot in general be expected 
to be equal to $\pth(\vec{z}|\vec{x})$, but in many cases is a reasonable approximation. 
We use RNNs due to their ability to scale well to large datasets. 

Table \ref{tab:recognition_models} details
the different choices for inference networks that we evaluate.
The Deep Kalman Smoother \textbf{\DKS} corresponds exactly to the functional form suggested by Eq. \ref{thm:p_fact}, and is our proposed variational approximation. The \DKS smoothes 
information from the past ($z_t$) and future ($x_t, \ldots x_{T}$) to form the approximate posterior distribution. 

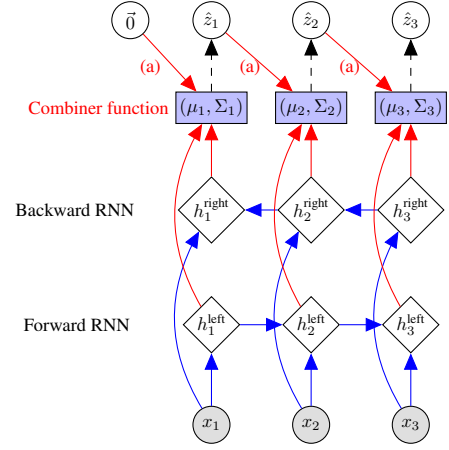
\begin{figure}[t]
  \centering 
	\begin{tikzpicture}[scale=.7, transform shape]
	\tikzstyle{recnet}=[rectangle,fill=blue!25,draw=black,minimum size=17pt,inner sep=2pt]
	\node [obs] (x1) {$x_1$};
	\node [obs, right= of x1, xshift=5pt] (x2) {$x_2$};
	\node [obs, right= of x2, xshift=5pt] (xT) {$x_3$};
		
	\node [det, above= of x1] (h1R) {$h_1^{\text{left}}$};
	\node [det, above= of x2] (h2R) {$h_2^{\text{left}}$};
	\node [det, above= of xT] (hTR) {$h_3^{\text{left}}$};
	\node [const, left= of h1R](hRlabel){Forward RNN};

	\node [det, above= of h1R] (h1L) {$h_1^{\text{right}}$};
	\node [det, above= of h2R] (h2L) {$h_2^{\text{right}}$};
	\node [det, above= of hTR] (hTL) {$h_3^{\text{right}}$};
	\node [const, left= of h1L, xshift=5pt](hRlabel){Backward RNN};

	\node [recnet, above= of h1L] (z1) {$(\mu_1, \Sigma_1)$};
	\node [recnet, above= of h2L] (z2) {$(\mu_2, \Sigma_2)$};
	\node [recnet, above= of hTL] (zT) {$(\mu_3, \Sigma_3)$};
	\node [const, left=of z1, xshift=25pt](comb){\textcolor{red}{Combiner function}};
	
	\node [const, above= of z1, yshift=-20pt, xshift=-33pt, red](label1){(a)};
	\node [const, above= of z2, yshift=-20pt, xshift=-33pt, red](label2){(a)};
	\node [const, above= of zT, yshift=-20pt, xshift=-33pt, red](label3){(a)};

	\node [latent, above= of z1] (zhat1) {$\hat{z}_1$};
	\node [latent, above= of z2] (zhat2) {$\hat{z}_2$};
	\node [latent, above= of zT] (zhatT) {$\hat{z}_3$};
   	\node [latent, left= of zhat1, xshift=5pt](zhat0) {$\vec{0}$};
   		
	\edge [blue]{h1R} {h2R};
	\edge [blue]{h2R} {hTR};
	
	\edge [blue]{h2L} {h1L};	
	\edge [blue]{hTL}{h2L};	
	
	\edge [blue]{x1}{h1R};
	\edge [blue]{x2} {h2R};
	\edge [blue]{xT} {hTR};
	\edge [blue, bend left]{x1} {h1L};
	\edge [blue, bend left]{x2} {h2L};
	\edge [blue, bend left]{xT} {hTL};
	\edge [red, bend left]{h1R} {z1};
	\edge [red, bend left]{h2R} {z2};
	\edge [red, bend left]{hTR} {zT};
	\edge [red]{h1L} {z1};
	\edge [red]{h2L} {z2};
	\edge [red]{hTL} {zT};
	\edge [dashed]{z1}{zhat1};
	\edge [dashed]{z2}{zhat2};
    \edge [dashed]{zT}{zhatT};
    \edge [red, shorten <=0pt]{zhat0}{z1};
    \edge [red]{zhat1}{z2};
    \edge [red]{zhat2}{zT};
	\end{tikzpicture} 
	\caption{\label{fig:var_approx_all}\small \textbf{Structured Inference Networks: } \textbf{MF-LR} and
          \textbf{ST-LR} variational approximations for a sequence of
          length $3$, using a bi-directional recurrent neural net
          (BRNN). The BRNN takes as input the sequence $(x_1, \ldots
          x_3)$, and through a series of non-linearities denoted by
          the \textcolor{blue}{blue} arrows it forms a sequence of
          hidden states summarizing information from the left and right 
          ($h_t^{\text{left}}$ and $h_t^{\text{right}}$)
          respectively. Then through a further sequence of
          non-linearities which we call the ``combiner function'' (marked (a) above), and
          denoted by the \textcolor{red}{red} arrows, it outputs two
          vectors $\mu$ and $\Sigma$, parameterizing the mean and
          diagonal covariance of $q_\phi(z_t|z_{t-1}, \vecx)$ of Eq.~\ref{eqn:q_fact}. Samples $\hat{z}_t$ are drawn from $q_\phi(z_t|z_{t-1},\vecx)$, as indicated by the black dashed arrows.
	For the structured variational models \textbf{ST-LR}, the samples $\hat{z}_t$ are fed into the computation of $\mu_{t+1}$ and $\Sigma_{t+1}$, as indicated by the red arrows with the label (a). The mean-field model does \emph{not} have these arrows, and therefore computes $q_\phi(z_t|\vecx)$. 
	We use $\hat{z}_0 = \vec{0}$.
	The inference network for \DKS (ST-R) is structured like that of ST-LR except without the RNN from the past. 
} 
\end{figure}

We also evaluate other possibilities for the variational models (inference
networks) $\qph$: two are mean-field models (denoted \textbf{MF}) and
two are structured models (denoted \textbf{ST}). They are
distinguished by whether they use information from the past (denoted
\textbf{L}, for left), the future (denoted \textbf{R}, for right), or
both (denoted \textbf{LR}). See Fig. \ref{fig:var_approx_all} for an
illustration of two of these methods.
Each conditions on a different subset of the observations
to summarize information in the input sequence $\vecx$. 
\DKS corresponds to \textbf{ST-R}. 

\begin{table}[t]
\centering 
\caption{\small \textbf{Inference Networks: } BRNN refers to a Bidirectional RNN and comb.fxn is shorthand for combiner function.}
\resizebox{\linewidth}{!}{
\begin{tabular}{ccc}
\toprule 
Inference Network & Variational Approximation for $z_t$ & Implemented With \\
\midrule 
\textbf{MF-LR} & $q(z_t|x_1, \ldots x_T)$ & BRNN\\
\textbf{MF-L} & $q(z_t|x_1,\ldots x_t)$ &  RNN\\
\textbf{ST-L} & $q(z_t|z_{t-1},x_1, \ldots x_t)$ &  RNN \& comb.fxn\\
\textbf{\DKS} & $q(z_t|z_{t-1},x_t, \ldots x_T)$ &  RNN \& comb.fxn \\
\textbf{ST-LR} & $q(z_t|z_{t-1},x_1, \ldots x_T)$ & BRNN \& comb.fxn \\
\bottomrule 
\end{tabular}\label{tab:recognition_models}
}
\end{table}

The hidden states of the RNN parameterize the variational
distribution, which go through what we call the ``combiner function''.
We obtain the mean $\mu_t$ and diagonal covariance $\sigma^2_t$ for the
approximate posterior at each time-step in a manner akin to Gaussian belief
propagation. Specifically, we interpret the hidden states of the
forward and backward RNNs as parameterizing the mean and variance of
two Gaussian-distributed ``messages'' summarizing the observations from the
past and the future, respectively. We then multiply these two
Gaussians, performing a variance-weighted average of the means. 
All operations should be understood to be performed
element-wise on the corresponding vectors.
$h_t^{\text{left}},h_t^{\text{right}}$ are the hidden states of the RNNs that run from the past and the future
respectively (see Fig. \ref{fig:var_approx_all}). 

{\bf Combiner Function for Mean Field Approximations:}
For the \textbf{MF-LR} inference network, the mean $\mu_t$ and
diagonal variances $\sigma^2_t$ of the variational distribution
$q_\phi(z_t|\vecx)$ are predicted using the output of the RNN (not
conditioned on $z_{t-1}$) as follows, where $\Softplus(x) = \log(1+\exp(x))$:
\begin{align}
	&\mu_{\text{r}}     = W^{\text{right}}_{\mu_{\text{r}}} h_t^{\text{right}} + b^{\text{right}}_{\mu_{\text{r}}}; \nonumber\\ 
	&\sigma^2_{\text{r}}= \Softplus(W^{\text{right}}_{\sigma^2_{\text{r}}} h_t^{\text{right}} + b^{\text{right}}_{\sigma^2_{\text{r}}})  \nonumber\\
	&\mu_{\text{l}}     = W^{\text{left}}_{\mu_{\text{l}}} h_t^{\text{left}} + b^{\text{left}}_{\mu_{\text{l}}};\nonumber\\
	&\sigma^2_{\text{l}}= \Softplus(W^{\text{left}}_{\sigma^2_{\text{l}}}h_t^{\text{left}} + b^{\text{left}}_{\sigma^2_{\text{l}}})  \nonumber\\
	&\mu_{t}           = \frac{\mu_{\text{r}}\sigma^2_{\text{l}}+\mu_{\text{l}}\sigma^2_{\text{r}}}{\sigma^2_{\text{r}}+\sigma^2_{\text{l}} }; \;\;
	\sigma^2_{t}       = \frac{\sigma^2_{\text{r}}\sigma^2_{\text{l}}}{\sigma^2_{\text{r}}+\sigma^2_{\text{l}}} \qquad\nonumber 
\end{align}

{\bf Combiner Function for Structured Approximations:} 
The combiner functions for the structured approximations are
implemented as:
\begin{align*}
&\mathtext{For {\bf ST-LR}} \nonumber\\
&h_{\text{combined}}= \frac{1}{3}(\Tanh(W z_{t-1}+b) + h_t^{\text{left}} + h_t^{\text{right}})\\ 
&\mathtext{For  {\bf \DKS}}\nonumber\\ 
&h_{\text{combined}}= \frac{1}{2}(\Tanh(W z_{t-1}+b) + h_t^{\text{right}})\\
&\mathtext{Posterior Means and Covariances} \;\;\\
&\mu_{t} = W_{\mu}h_{\text{combined}}+b_{\mu} \nonumber\\
&\sigma^2_{t} = \Softplus(W_{\sigma^2}h_{\text{combined}}+b_{\sigma^2}) 
\end{align*}
The combiner function uses the $\Tanh$ non-linearity from $z_{t-1}$ to
approximate the transition function (alternatively, one could share
parameters with the generative model), and here we use a
simple weighting between the components.

\textbf{Relationship to Related Work: }
\citeauthor{archer2015black,gao2016linear} use $q(\vec{z}|\vec{x}) = \prod_t q(z_t|z_{t-1},\vec{x})$ where $q(z_t|z_{t-1},\vec{x}) = \mathcal{N}\left(\mu(x_t),\Sigma(z_{t-1},x_t,x_{t-1})\right)$. 
The key difference from our approach is that this parameterization (in particular, conditioning the posterior means only on $x_t$) does not account for the information
from the future relevant to the approximate posterior distribution for $z_t$. 

\citeauthor{johnson2016structured} 
interleave predicting the local variational parameters of the graphical model (using an inference network) with steps of message passing inference.
A key difference between our approach and theirs is that we rely on the structured inference network to predict the optimal local variational parameters
directly. In contrast, in \citeauthor{johnson2016structured}, any suboptimalities 
in the initial local variational parameters may be overcome by the subsequent steps of optimization 
albeit at additional computational cost. 

\citeauthor{chung2015recurrent} propose the Variational RNN (VRNN) in
which Gaussian noise is introduced at each time-step of a RNN.
\citeauthor{chung2015recurrent} use an inference network that shares
parameters with the generative model and only uses information from
the past. If one views the noise variables and the hidden state of the
RNN at time-step $t$ together as $z_t$, then a factorization similar
to Eq. \ref{eqn:bound_likelihood} can be shown to hold, although the
KL term would no longer have an analytic form since $\pth(z_t|z_{t-1},
x_{t-1})$ would not be Normally distributed. Nonetheless, our same
structured inference networks (i.e. using an RNN to summarize
observations from the future) could be used to improve the tightness
of the variational lower bound, and our empirical results suggest that
it would result in better learned models.

\section{Deep Markov Models}
Following \cite{Raiko2006}, we apply the ideas of deep learning to non-linear continuous state space models.
When the transition and emission function have an unknown functional form, we 
parameterize $\meanfxn,\covfxn,\emisfxn$ from Eq. \ref{eqn:gen_model}
with deep neural networks. See Fig. \ref{fig:dkf} (right) for an illustration of the graphical model.  

{\bf Emission Function:} 
We parameterize the emission function $\emisfxn$ using a two-layer MLP (multi-layer perceptron),
$\text{MLP}(x, \text{NL}_1, \text{NL}_2) = \text{NL}_2 (W_2 \text{NL}_1(W_1x+b_1)+b_2))$,
where NL denotes non-linearities such as ReLU, sigmoid, or
tanh units applied element-wise to the input vector. 
For modeling binary data, $\emisfxn(z_t)
=\Sigmoid(W_{\text{emission}}\MLP(z_t,\ReLU,\ReLU)+b_{\text{emission}})$
parameterizes the mean probabilities of independent Bernoullis. 

{\bf Gated Transition Function:} 
We parameterize the transition function from $z_t$ to $z_{t+1}$ using
a gated transition function inspired by Gated
Recurrent Units \citep{chung2014empirical}, instead of an MLP.
Gated recurrent units (GRUs) are a neural architecture that parameterizes the recurrence equation in the 
RNN with gating units to control the flow of information from one hidden state to the next, conditioned on the observation. 
Unlike GRUs, in the \DKF, the transition function is not conditional on any of the observations. All the information
must be encoded in the completely stochastic latent state. To achieve this goal, we create 
a Gated Transition Function. 
We would like the model to have the flexibility to choose
a linear transition for some dimensions while having
a non-linear transitions for the others. We 
adopt the following parameterization, where $\mathbb{I}$ denotes 
the identity function and $\odot$ denotes element-wise multiplication:
\begin{align*}
&g_t = \MLP(z_{t-1},\ReLU,\Sigmoid) \;\; \mathtext{Gating Unit}\\
&h_t = \MLP(z_{t-1},\ReLU,\mathbb{I}) \;\; \mathtext{Proposed mean}\nonumber\\
&\mathtext{Transition Mean $\meanfxn$ and $\covfxn$}\nonumber\\
&\mu_t(z_{t-1})  = (1-g_t)\odot(W_{\mu_p}z_{t-1}+b_{\mu_p}) +g_t\odot h_t \nonumber\\
&\sigma^2_t(z_{t-1}) = \Softplus( W_{\sigma_p^2} \ReLU(h_t)+ b_{\sigma_p^2})\nonumber
\end{align*}

Note that the mean and covariance functions both share the use of $h_t$.
In our experiments, we initialize $W_{\mu_p}$ to be the identity function and $b_{\mu_p}$ to $0$.
The parameters of the emission and transition function 
form the set $\theta$. 

\section{Evaluation}
Our models and learning algorithm are implemented in Theano \cite{theano}. 
We use Adam \cite{kingma2014adam} with a learning rate of $0.0008$ to train the \DKF. 
Our code is available at \texttt{\small github.com/clinicalml/structuredinference}.

\textbf{Datasets: } We evaluate on three datasets. 

\textit{Synthetic: } We consider simple linear and non-linear GSSMs. To train
the inference networks we use $N=5000$ datapoints of length $T=25$. We consider
both one and two dimensional systems for inference and parameter estimation. 
We compare our results using the training value of the variational bound $\lthph$ (Eq. \ref{eqn:bound_likelihood}) and the
$\text{RMSE} = \sqrt{\frac{1}{N} \frac{1}{T}\sum_{i=1}^N \sum_{t=1}^T[\mu_{\phi}(x_{i,t})-z^*_{i,t}]^2}$,
where $z^*$ correspond to the true underlying $z$'s that generated the data. 

\textit{Polyphonic Music: }
We train DMMs on polyphonic music data \cite{boulanger2012modeling}. 
An instance in the sequence comprises 
an 88-dimensional binary vector corresponding to the notes of a piano. 
We learn for $2000$ epochs and report results based on early stopping using the validation set. 
We report held-out negative log-likelihood (NLL) in the format ``a (b) \{c\}''.
$a$ is an importance sampling based estimate of the
NLL (details in supplementary material);
$b = \frac{1}{\sum_{i=1}^N T_i}\sum_{i=1}^N -\mathcal{L}(\vecx;\theta, \phi)$ where $T_i$ is the length
of sequence $i$. This is an upper bound on the NLL, which facilitates comparison
to RNNs;
TSBN \cite{gan2015deep} (in their code) report 
$c = \frac{1}{N}\sum_{i=1}^N \frac{1}{T_i} \mathcal{L}(\vecx;\theta, \phi)$. We compute this 
to facilitate comparison with their work.
As in \citep{howtotrain}, we found 
annealing the KL divergence in the variational bound ($\lthph$) from $0$ to $1$ over
$5000$ parameter updates got better results. 

\textit{Electronic Health Records (EHRs): } 
The dataset comprises $~5000$ diabetic patients using data from a major health insurance provider. 
The observations of interest are: A1c level (hemoglobin A1c, a protein
for which a high level indicates that the patient is diabetic) and glucose (blood sugar).
We bin glucose into quantiles and A1c into clinically meaningful bins. The observations also include age, gender and ICD-9 
diagnosis codes for co-morbidities of diabetes such as congestive heart failure, chronic kidney disease and obesity. 
There are $48$ binary observations for a patient at every time-step. 
We group each patient's data (over $4$ years) into three month intervals, yielding a sequence of length $18$. 

\subsection{Synthetic Data}
\textbf{Compiling Exact Inference: }
We seek to understand whether inference networks can accurately 
compile exact posterior inference into the network parameters $\phi$
for linear GSSMs when exact inference is feasible.
For this experiment we optimize Eq. \ref{eqn:bound_likelihood} over
$\phi$, while $\theta$ is fixed to a synthetic distribution given by a
one-dimensional GSSM. 
We compare results obtained by the various approximations we propose 
to those obtained by an implementation of Kalman smoothing \citep{kfimplementation} which performs \emph{exact inference}. 
Fig. \ref{fig:synthetic_linear} (top and middle) depicts our results. 
The proposed \textbf{\DKS} (i.e., {\textbf{ST-R}) and \textbf{ST-LR}
outperform the mean-field based variational method \textbf{MF-L} that
only looks at information from the past. \textbf{MF-LR}, however, is
often able to catch up when it comes to RMSE,
highlighting the role that information from the 
future plays when performing posterior 
inference, as is evident in the posterior factorization in Eq. \ref{thm:p_fact}. 
Both \textbf{\DKS} and \textbf{ST-LR} converge to the RMSE of the
exact Smoothed KF, and moreover their lower bound on the likelihood becomes tight.

\begin{figure}[t!]
	\begin{center}
		\includegraphics[width=0.45\textwidth,keepaspectratio]{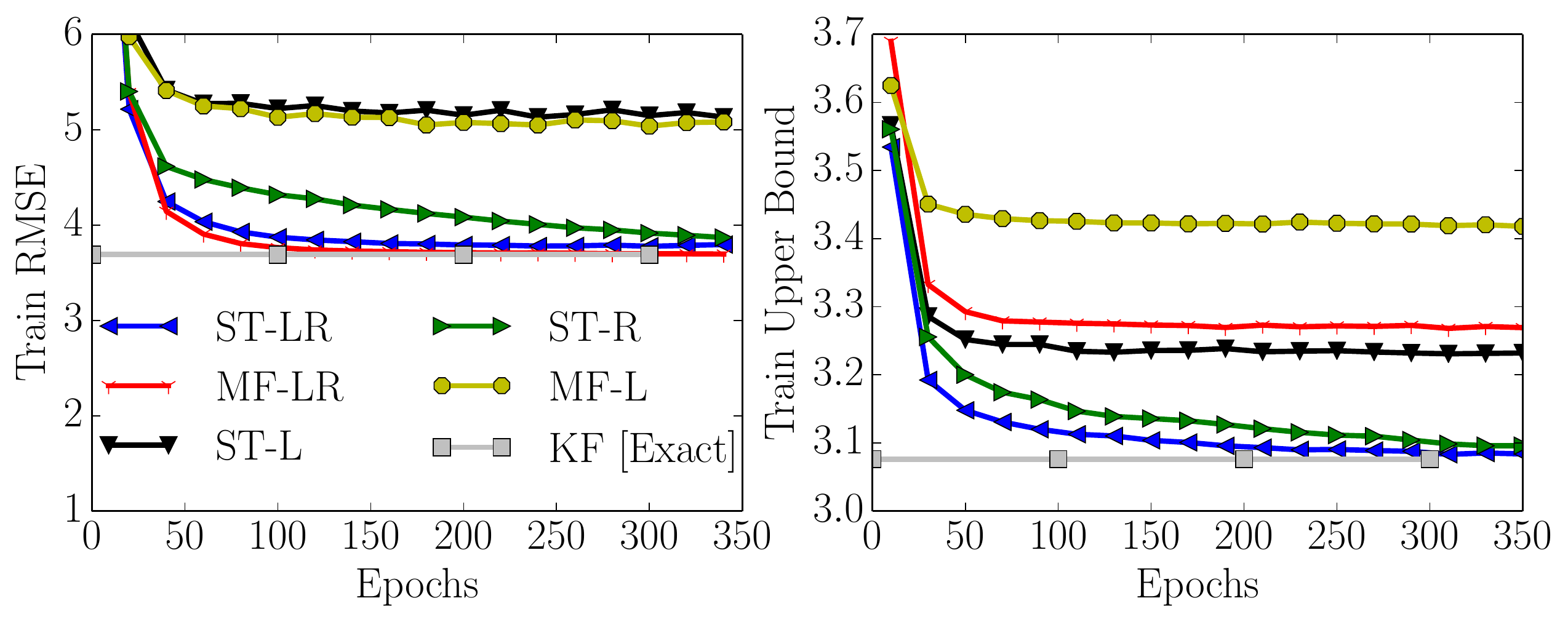}
		\includegraphics[width=0.45\textwidth,keepaspectratio]{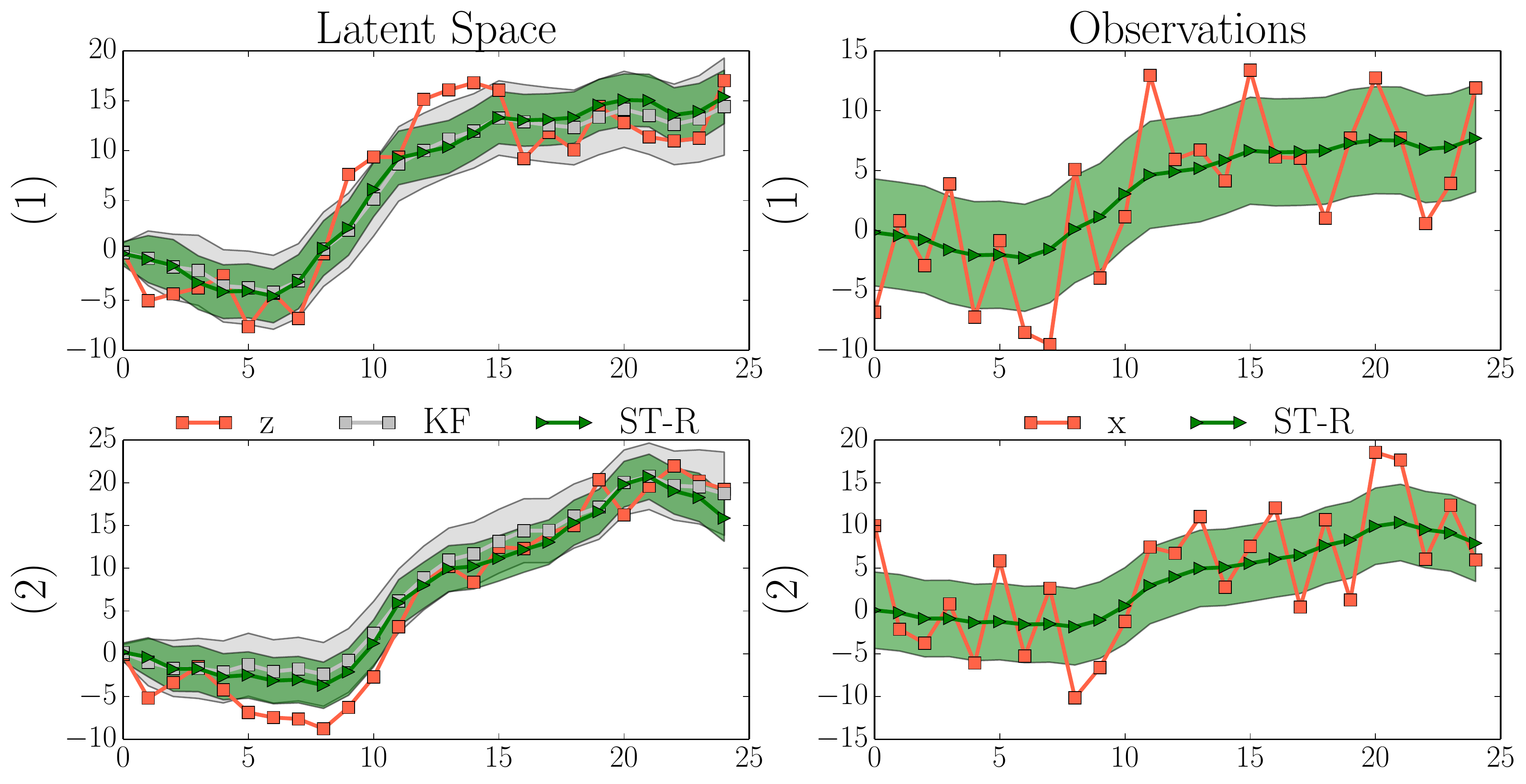}
	\end{center}
\caption{\label{fig:synthetic_linear}
\small \textbf{Synthetic Evaluation: } ({\bf Top \& Middle}) Compiled inference for a \emph{fixed} linear
  GSSM: $z_t \sim \mathcal{N}(z_{t-1} + 0.05, 10)$, $x_t \sim \mathcal{N}(0.5z_t, 20)$.
The training set comprised $N=5000$ one-dimensional 
observations of sequence length $T=25$.
{\bf (Top left)} RMSE with respect to true $z^*$
  that generated the data. 
{\bf (Top right)} Variational bound during
  training. The results on held-out data are very similar (see supplementary material).
{\bf (Bottom)} Visualizing inference in two sequences (denoted (1) and (2)); Left panels show the Latent Space of variables $z$, right panels show the Observations $x$. Observations are generated by the application of the 
emission function to the posterior shown in Latent Space.
Shading denotes standard deviations.
}
\end{figure}

\begin{figure}[t!]
	\centering
	\includegraphics[width=0.45\textwidth,keepaspectratio]{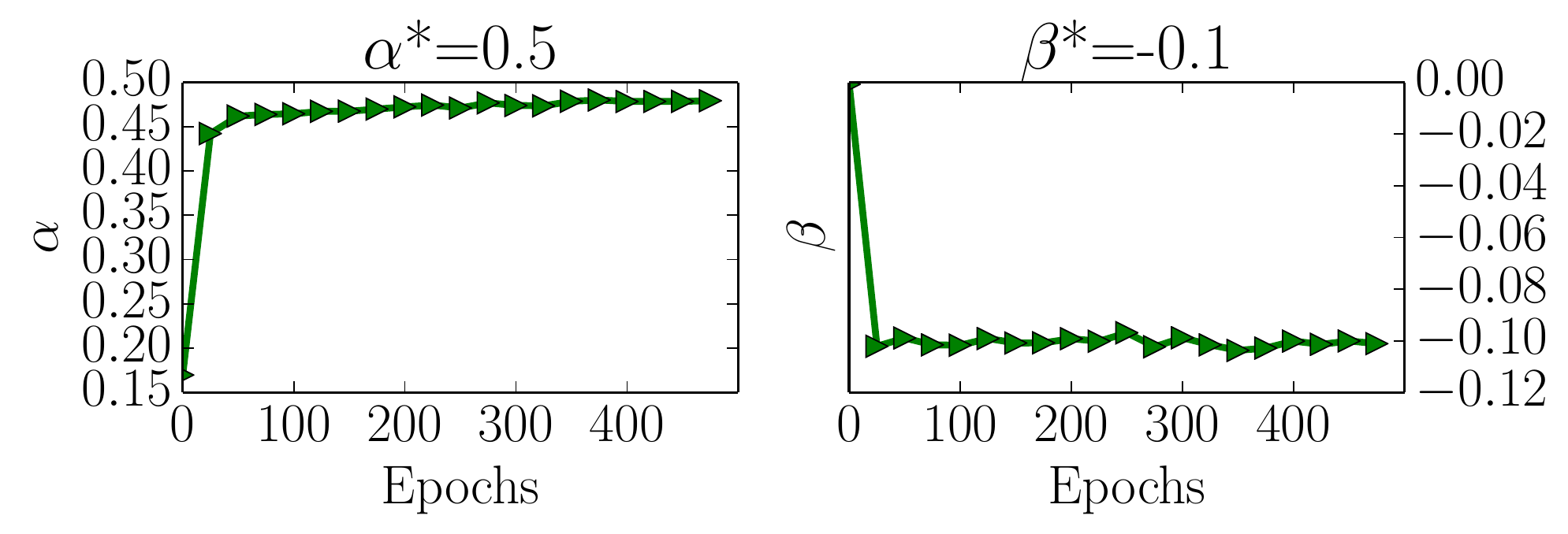}
\caption{\label{fig:synthetic_param_est}
\small 
\textbf{Parameter Estimation: }
Learning parameters $\alpha,\beta$ in a two-dimensional non-linear GSSM. $N=5000, T=25$
$\vec{z}_t\sim\mathcal{N}([0.2z_{t-1}^0+\text{tanh}(\alpha z_{t-1}^1); 0.2z_{t-1}^1+\sin(\beta z_{t-1}^0)] ,1.0)$
$\vec{x}_t\sim\mathcal{N}(0.5\vec{z}_t,0.1)$ where $\vec{z}$ denotes a vector, $[]$ denotes concatenation and superscript denotes
indexing.
}
\end{figure}

\textbf{Approximate Inference and Parameter Estimation: }
Here, we experiment with applying the inference networks to synthetic non-linear generative models
as well as using \DKS for learning a subset of parameters within a fixed generative model.
On synthetic non-linear datasets (see supplemental material)
we find, similarly, that the structured variational approximations 
are capable of matching the performance of inference using a smoothed Unscented Kalman Filter \cite{wan2000unscented} on held-out data. 
Finally, Fig. \ref{fig:synthetic_param_est} illustrates a toy
instance where we successfully perform 
parameter estimation in a synthetic, two-dimensional, non-linear GSSM.

\subsection{Polyphonic Music}
\textbf{Mean-Field vs Structured Inference Networks: }
Table \ref{tab:polyphonic_structural} shows the
results of learning a \DMM on the polyphonic music dataset using 
\textbf{MF-LR}, \textbf{ST-L}, \textbf{\DKS} and \textbf{ST-LR}.
\textbf{ST-L} is a structured variational approximation that only considers
information from the past and, up to implementation details, is comparable
to the one used in \cite{gregor2015draw}.
Comparing the negative log-likelihoods of the learned models, we see
that the looseness in the variational bound (which we first observed in the synthetic
setting in Fig. \ref{fig:synthetic_linear} top right) significantly
affects the ability to learn. \textbf{ST-LR} and \textbf{\DKS}
substantially outperform \textbf{MF-LR} and \textbf{ST-L}.
This adds credence to 
the idea that by taking into consideration the factorization of the posterior, one can perform
better inference and, consequently, learning, in real-world, high
dimensional settings. Note that the \textbf{\DKS} network has half the parameters
of the \textbf{ST-LR} and \textbf{MF-LR} networks.

\textbf{A Generalization of the \DMM: }
To display the efficacy of our inference algorithm to model variants beyond
first-order Markov Models, we further augment the \DKF with edges from $x_{t-1}$ to $z_t$
and from $x_{t-1}$ to $x_t$.  
We refer to the resulting generative model as \DKF-Augmented (Aug.).
Augmenting the \DKF
with additional edges 
realizes a richer
class of generative models.

We show that \textbf{\DKS} can be used \emph{as is} for inference on 
a more complex generative model than \DMM, while making gains in held-out likelihood. All following experiments use {\bf \DKS} for posterior inference.

The baselines we compare to in Table \ref{tab:polyphonic} also have more complex generative models than the \DKF. 
STORN has edges from $x_{t-1}$ to $z_{t}$ given by the recurrence update
and TSBN has edges from $x_{t-1}$ to $z_{t}$ as well as from $x_{t-1}$ to $x_t$.  
HMSBN shares
the same structural properties as the \DKF, but
is learned using a simpler inference network. 

In Table~\ref{tab:polyphonic}, 
as we increase the complexity of the generative model, we obtain better results across all datasets. 

The \DKF outperforms both RNNs and HMSBN everywhere, outperforms STORN on JSB, Nottingham 
and outperform TSBN on all datasets except Piano. 
Compared to LV-RNN (that optimizes the inclusive KL-divergence), 
\DKF-Aug obtains better results on all datasets except JSB. %
This showcases our flexible, structured inference network's ability to learn 
powerful generative models that compare favourably to other 
state of the art models. We provide audio
files for samples from the learned \DMM models in the code repository. 

\begin{table}[t]
	\centering
	\caption{\small \textbf{Comparing Inference Networks: }  Test negative log-likelihood on polyphonic music of different inference networks trained
	on a \DKF with a fixed structure (lower is better). The numbers
      inside parentheses are the variational bound.}
\resizebox{\linewidth}{!}{
\begin{tabular}{ccccc}
    \toprule
    ~Inference Network & JSB  & Nottingham & Piano & Musedata\\
    \midrule 
    {\bf \DKS} (i.e., {\bf ST-R})&  \begin{tabular}{c} 6.605 (7.033)  \end{tabular} 
    & \begin{tabular}{c} 3.136 (3.327) \end{tabular}
    & \begin{tabular}{c} 8.471 (8.584)\end{tabular}
    & \begin{tabular}{c} 7.280 (7.136)\end{tabular}\\
    \midrule
    {\bf ST-L}&  \begin{tabular}{c} 7.020 (7.519)  \end{tabular} 
    & \begin{tabular}{c} 3.446 (3.657) \end{tabular}
    & \begin{tabular}{c} 9.375 (9.498)\end{tabular}
    & \begin{tabular}{c} 8.301 (8.495)\end{tabular}\\
    \midrule
    {\bf ST-LR }&  \begin{tabular}{c} 6.632 (7.078)  \end{tabular} 
    & \begin{tabular}{c}  3.251 (3.449) \end{tabular}
    & \begin{tabular}{c}  8.406 (8.529)\end{tabular}
    & \begin{tabular}{c}  7.127 (7.268)\end{tabular}\\
    \midrule
    {\bf MF-LR}&  \begin{tabular}{c} 6.701  (7.101) \end{tabular} 
    & \begin{tabular}{c} 3.273 (3.441)  \end{tabular}
    & \begin{tabular}{c} 9.188 (9.297) \\ \end{tabular}
    & \begin{tabular}{c} 8.760 (8.877) \end{tabular}\\
    \bottomrule
    \end{tabular}\label{tab:polyphonic_structural}
   }
\end{table}
 
\begin{table}[t]
	\centering
	\caption{
		\label{tab:polyphonic}
		\small{\textbf{Evaluation against Baselines: } Test negative log-likelihood (lower is better) on Polyphonic Music Generation dataset.
			\textbf{Table Legend}: RNN \cite{boulanger2012modeling}, 
			LV-RNN \cite{gu2015neural}, STORN \cite{bayer2014learning},
			TSBN, HMSBN \cite{gan2015deep}.}}
\resizebox{\linewidth}{!}{
\begin{tabular}{ccccc}
    \toprule
    ~Methods       & JSB  & Nottingham & Piano & Musedata\\
    \midrule
    \midrule
    \DKF &  \begin{tabular}{c} 6.388 \\(6.926)\\ \{6.856\}\end{tabular} 
    & \begin{tabular}{c} 2.770 \\(2.964)\\ \{2.954\} \end{tabular}
    & \begin{tabular}{c}  7.835 \\(7.980)\\ \{8.246\} \end{tabular}
    & \begin{tabular}{c} 6.831 \\(6.989)\\ \{6.203\} \end{tabular}\\
    \midrule
    \DKF-Aug. &  \begin{tabular}{c} 6.288 \\(6.773)\\ \{6.692\}\end{tabular} 
    & \begin{tabular}{c} 2.679\\(2.856)\\ \{2.872\} \end{tabular}
    & \begin{tabular}{c} 7.591\\(7.721)\\ \{8.025\} \end{tabular}
    & \begin{tabular}{c} 6.356\\(6.476)\\ \{5.766\} \end{tabular}\\
    \midrule
    HMSBN &  \begin{tabular}{c}(8.0473)\\ \{7.9970\}\end{tabular} 
    & \begin{tabular}{c} (5.2354)\\ \{5.1231\} \end{tabular}
    & \begin{tabular}{c} (9.563)\\ \{9.786\}  \end{tabular}
    & \begin{tabular}{c} (9.741)\\ \{8.9012\}  \end{tabular}\\
    \midrule
    STORN& 6.91 & 2.85 & 7.13 & 6.16\\
    \midrule
    RNN  & 8.71 & 4.46 & 8.37 & 8.13\\ 
    \midrule
    TSBN & \{7.48\} & \{3.67\} & \{7.98\} & \{6.81\} \\
    \midrule
    LV-RNN & 3.99 & 2.72 & 7.61 & 6.89\\
    \bottomrule
    \end{tabular}
}
\end{table}

\subsection{EHR Patient Data}
Learning models from large observational
health datasets is a promising approach to advancing
precision medicine and could be used, for example, to understand which
medications work best, for whom. In this section, 
we show how a \DMM may be used for precisely such an application.
Working with EHR data poses some technical challenges: EHR data are noisy, high dimensional and difficult to characterize easily.
Patient data is rarely contiguous over large parts of the dataset and is often missing (not at random). We learn a \DMM on the data 
showing how to handle the aforementioned technical challenges and use it for model based counterfactual prediction. 

\textbf{Graphical Model: }
Fig. \ref{fig:dmm_action} represents the generative model we use when $T=4$. 
The model captures the idea of an underlying time-evolving latent state for a patient ($z_t$) that
is solely responsible for the diagnosis codes and lab values ($x_t$) we observe. In addition, the patient state
is modulated by drugs ($u_t$) prescribed by the doctor. We may assume that the drugs prescribed at any point in time
depend on the patient's entire medical history
though in practice, the dotted edges in the Bayesian network never need to be modeled since $x_t$ and $u_t$ are always assumed to be observed. A natural line of follow up work would be to consider learning when $u_t$ is missing or latent.  

We make use of time-varying (binary) drug prescription $u_t$ for each patient
by augmenting the \DMM with an additional edge every time step. 
Specifically, the \DMM's transition function is now 
$z_t\sim\mathcal{N}(\meanfxn(z_{t-1},u_{t-1}),\covfxn(z_{t-1},u_{t-1}))$ (cf. Eq. \ref{eqn:gen_model}).
In our data, each $u_t$ is an indicator vector of eight anti-diabetic drugs including Metformin and Insulin, where Metformin is the most commonly prescribed first-line anti-diabetic drug.

\begin{figure}[h]
\centering
        \begin{tikzpicture}[scale=0.7, transform shape]
        \node[latent] (z1) {$z_1$};
        \node[obs, above= of z1] (u1) {$u_1$};
        \node[obs,below= of z1] (x1) {$x_1$};
        \node[latent,right=of z1] (z2) {$z_2$};
        \node[obs, above=of z2] (u2) {$u_2$};
        \node[obs, below=of z2] (x2) {$x_2$};
        \node[latent,right=of z2] (z3) {$z_3$};
        \node[obs, above=of z3] (u3) {$u_3$};
        \node[obs, below=of z3] (x3) {$x_3$};
        \node[latent,right=of z3] (z4) {$z_4$};
        \node[obs, below=of z4] (x4) {$x_4$};
	\edge{z1}{z2};
	\edge{z2}{z3};
	\edge{z3}{z4};
	\edge{z1}{x1};
	\edge{z2}{x2};
	\edge{z3}{x3};
	\edge{z4}{x4};
	\edge{u1}{z2};
	\edge{u2}{z3};
	\edge{u3}{z4};
	\edge[dashed] {x1}{u2};
	\edge[dashed] {x1}{u3};
	\edge[dashed] {x2}{u3};
    \end{tikzpicture}
    \caption{\small \textbf{\DMM for Medical Data: } 
    The \DMM (from Fig. \ref{fig:dkf}) is augmented with external actions $u_t$ representing medications presented to the patient. $z_t$ is 
the latent state of the patient. $x_t$ are the observations that we model.
Since both $u_t$ and $x_t$ are always assumed observed, the conditional distribution $p(u_t|x_{1},\ldots,x_{t-1})$ 
may be ignored during learning.}
\label{fig:dmm_action}
\end{figure}
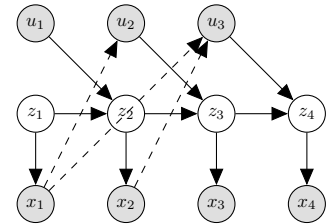

\textbf{Emission \& Transition Function:}The choice of emission and transition function to use for 
such data is not well understood. In Fig. \ref{fig:cfac} (right), we experiment with variants of 
DMMs and find that using MLPs (rather than linear functions) in the emission and transition function yield the best generative models in terms of held-out likelihood. 
In these experiments, the hidden dimension was set as $200$ for the emission and transition functions.
We used an RNN size of $400$ and a latent dimension of size $50$. We use the \DKS as our inference network for learning.  

\textbf{Learning with Missing Data: }
In the EHR dataset, a subset of the observations (such as A1C and Glucose
values which are commonly used to assess blood-sugar levels for diabetics) is frequently missing in the data. We marginalize them 
out during learning, which is straightforward within the probabilistic semantics of our Bayesian network. 
The sub-network of the original graph we are concerned with is the emission function since missingness affects our ability 
to evaluate $\log p(x_t|z_t)$ (the first term in Eq. \ref{eqn:bound_likelihood}). 
The missing random variables are leaves in the Bayesian sub-network (comprised of the emission function).  
Consider a simple example of two modeling two observations
at time $t$, namely $m_t,o_t$. The log-likelihood of the data ($m_t,o_t$) conditioned 
on the latent variable $z_t$ decomposes as $\log p(m_t,o_t|z_t) = \log p(m_t|z_t)+ \log p(o_t|z_t)$
since the random variables are conditionally independent given their parent. 
If $m$ is missing and marginalized out while $o_t$ is observed, then
our log-likelihood is: $\log \int_m p(m_t, o_t|z_t)  = \log (\int_m p(m_t|z_t) p(o_t|z_t) ) = \log p(o_t|z_t)$ (since $\int_m p(m_t|z_t) = 1$) 
i.e we effectively ignore the missing observations when estimating the log-likelihood of the data. 

\begin{figure}[t!]
	\centering
	\includegraphics[width=\linewidth]{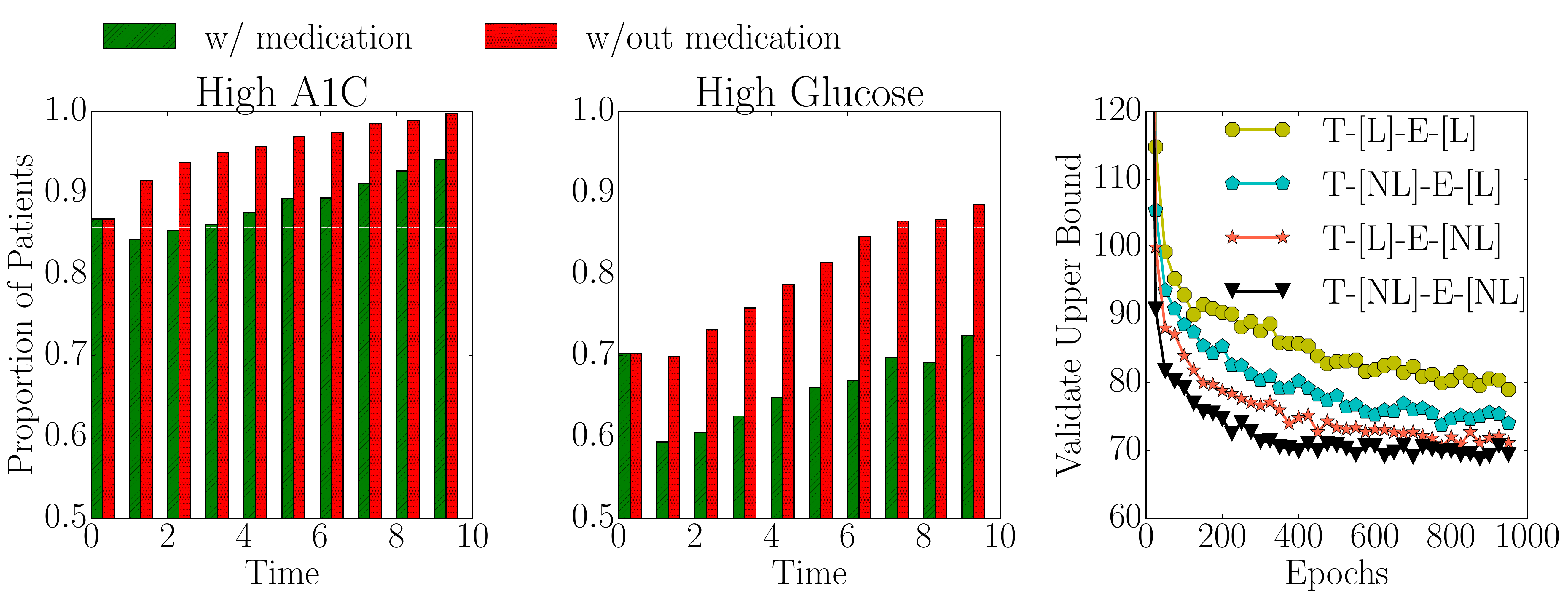}
	\caption{\small {\bf (Left Two Plots) } 
	Estimating Counterfactuals with \DKF:  The x-axis denotes the number of $3$-month intervals after prescription of Metformin. 
		The y-axis denotes the proportion of patients (out of a test set size of $800$) who, after their first prescription of Metformin, experienced
	a high level of A1C. In each tuple of bar plots at every time step, the left aligned bar plots (green) represent the population that received diabetes
medication while the right aligned bar plots (red) represent the population that did not receive diabetes medication.
{\bf (Rightmost Plot) } Upper bound on negative-log likelihood for different DMMs trained on the medical data. (T) denotes ``transition'', (E) denotes ``emission'', (L) denotes ``linear'' and (NL) denotes ``non-linear''. 
}
	\label{fig:cfac}
\end{figure}
\textbf{The Effect of Anti-Diabetic Medications: } 
Since our cohort comprises diabetic patients, we ask a counterfactual question: what \emph{would 
have happened} to a patient had anti-diabetic drugs not been prescribed? Specifically we are interested in the patient's blood-sugar level as measured by the widely-used A1C blood-test.
We perform inference using held-out patient data leading up to the time $k$ of first prescription of Metformin.
From the posterior mean, we perform ancestral sampling tracking two latent trajectories: 
(1) the factual: where we sample new latent states conditioned 
on the medication $u_t$ the patient had actually received and 
(2) the counterfactual: where we sample conditioned on 
not receiving any drugs for all remaining timesteps (i.e $u_k$ set to the zero-vector). 
We reconstruct the patient observations $x_k,\ldots,x_T$, threshold
the predicted values of A1C levels into high and low and visualize the average number of high A1C levels 
we observe among the synthetic patients in both scenarios. 
This is an example of performing do-calculus \cite{pearl2009causality} in order to estimate model-based counterfactual effects.

The results are shown in Fig. \ref{fig:cfac}. We see the model learns that, on average, patients who were prescribed anti-diabetic medication
had more controlled levels of A1C than patients who did not receive any medication.  
Despite being an aggregate effect, this is interesting because it is a phenomenon that coincides with our intuition
but was confirmed by the model in an entirely unsupervised manner. Note that in our dataset, most diabetic patients are indeed prescribed anti-diabetic medications, making the counterfactual prediction harder. The ability of this model to answer such
queries opens up possibilities into building personalized neural models of healthcare. 
Samples from the learned generative model
and implementation details may be found in the supplement.

\section{Discussion}

We introduce a general algorithm for scalable learning in a rich family of latent variable models for time-series data. The underlying methodological principle we propose is to build the inference network to mimic the posterior distribution (under the generative model).
The space complexity of our learning algorithm depends neither on the sequence
length $T$ nor on the training set size $N$, offering massive
savings compared to classical variational inference methods.

Here we propose and evaluate building variational inference networks to mimic
the structure of the true posterior distribution.  
Other structured variational approximations 
are also possible. %
For example, one could instead use an RNN 
from the past, conditioned on a summary statistic of the future,
during learning and inference.

Since we use RNNs only in the inference network, 
it should be possible to continue to increase their capacity and 
condition on different modalities that might be relevant to approximate 
posterior inference without worry of overfitting the data. 
Furthermore, this confers us the ability to easily model in the presence of missing data 
since the semantics of the \DMM render it easy to marginalize out unobserved data.
In contrast, in a (stochastic) RNN (bottom in Fig. \ref{fig:dkf}) it is much more difficult
to marginalize out unobserved data due to the dependence of the intermediate hidden states on the previous
input. Indeed this allowed us to develop a principled application of the learning algorithm to
modeling longitudinal patient data in EHR data and 
inferring treatment effect.  

\section*{Acknowledgements}
The Tesla K40s used for this research were donated by the NVIDIA 
Corporation. The authors gratefully acknowledge support by the DARPA Probabilistic Programming for Advancing
Machine Learning (PPAML) Program under AFRL prime contract
no. FA8750-14-C-0005, ONR \#N00014-13-1-0646, a NSF CAREER award
\#1350965, and Independence Blue Cross. We thank David Albers, Kyunghyun Cho, Yacine Jernite, Eduardo Sontag and anonymous reviewers
for their valuable feedback and comments.

\bibliographystyle{aaai}
\bibliography{refs}

\clearpage
\appendix
\centerline{\large \bf Appendix}
\section{Lower Bound on the Likelihood of data\label{appsec:lb_likelihood}}

We can derive the bound on the likelihood $\lthph$ as follows:
\begin{align}
	\label{eqn:bound_likelihood_app}
&\log \pth(\vecx) \geq \int_{\vecz} \qph(\vecz|\vecx) \log \frac{\pth(\vecz)\pth(\vecx|\vecz)}{\qph(\vecz|\vecx)} d\vecz \nonumber\\
&= \Exp{\log \pth(\vecx|\vecz)}{\qph(\vecz|\vecx)} - \KL(\qph(\vecz|\vecx)||\pth(\vecz))\nonumber\\
&\mathtext{ Using $x_{t}\bigCI x_{\neg t}|z_t$ } \nonumber \\
&= \sum_{t=1}^T\Exp{\log \pth(x_t|z_t)}{\qph(z_t|\vecx)} - \KL(\qph(\vecz|\vecx)||\pth(\vecz)) \\
&= \lthph \nonumber
\end{align}
In the following we omit the dependence of $q$ on $\vecx$, and omit the subscript $\phi$.
We can show that the $\KL$ divergence between the approximation to the posterior and the prior simplifies as:
\begin{align}
	&\KL(q(z_1,\ldots,z_T)||p(z_1,\ldots,z_T))\nonumber\\
	&= \int_{z_1}\ldots\int_{z_T} q(z_1)\ldots q(z_T|z_{T-1}) \log \frac{p(z_1,\ldots,z_T)}{q(z_1).. q(z_T|z_{T-1})}\nonumber\\
	&\mathtext{Factorization of the variational distribution}\nonumber\\ 
	&= \int_{z_1}\ldots\int_{z_T} q(z_1)\ldots q(z_T|z_{T-1})\nonumber\\
	&\log \frac{p(z_1) p(z_2|z_1)\ldots p(z_T|z_{T-1})}{q(z_1)\ldots q(z_T|z_{T-1})}\nonumber\\ 
	&\mathtext{Factorization of the prior}\nonumber\\
	&= \int_{z_1}\ldots\int_{z_T} q(z_1)\ldots q(z_T|z_{T-1}) \log \frac{p(z_1)}{q(z_1)}+\nonumber\\
	&\sum_{t=2}^{T} \int_{z_1}\ldots\int_{z_T} q(z_1)\ldots q(z_T|z_{T-1}) \log\frac{p(z_t|z_{t-1})}{q(z_t|z_{t-1})}\nonumber\\
	&= \int_{z_1} q(z_1)\log\frac{p(z_1)}{q(z_1)} +  \sum_{t=2}^{T} \int_{z_{t-1}}\int_{z_{t}}q(z_t)\log\frac{p(z_t|z_{t-1})}{q(z_t|z_{t-1})}\nonumber\\
	&\mathtext{Each expectation over $z_t$ is constant for $t\notin\{t,t-1\}$}\nonumber\\
	&= \KL(q(z_1)||p(z_1)) \nonumber\\
	&+ \sum_{t=2}^{T} \Exp{\KL(q(z_t|z_{t-1})||p(z_t|z_{t-1}))}{q(z_{t-1})}\nonumber\\
\end{align}

For evaluating the marginal likelihood on the test set, we can use the following Monte-Carlo estimate:
\begin{equation}
	p(\vecx)\approxeq \frac{1}{S} \sum_{s=1}^S \frac{p(\vecx|\vecz^{(s)}) p(\vecz^{(s)}) }{q(\vecz^{(s)}|\vecx)} \;\;\; \vecz^{(s)}\sim q(\vecz|\vecx)
\end{equation}
This may be derived in a manner akin to the one depicted in Appendix E \citep{rezende2014stochastic} or Appendix D \citep{kingma2013auto}.

The log likelihood on the test set is computed using: 
\begin{equation}
	\label{eqn:test_ll_logsum}
	\log p(\vecx)\approxeq \log \frac{1}{S} \sum_{s=1}^S \exp \log\left[\frac{p(\vecx|\vecz^{(s)}) p(\vecz^{(s)}) }{q(\vecz^{(s)}|\vecx)}\right]
\end{equation}
Eq. \ref{eqn:test_ll_logsum} may be computed in a numerically stable manner using the log-sum-exp trick.  

\section{KL divergence between Prior and Posterior\label{appsec:kldiv}}

Maximum likelihood learning requires us to compute:
\begin{align}
	\label{eqn:KLdiv_factorized_app}
&\KL(q(z_1,\ldots,z_T)||p(z_1,\ldots,z_T))\nonumber\\
&= \KL(q(z_1)||p(z_1)) \nonumber\\
&+ \sum_{t=2}^{T-1} \Exp{\KL(q(z_t|q_{t-1})||p(z_t|z_{t-1}))}{q(z_{t-1})}
\end{align}

The KL divergence between two multivariate Gaussians $q$, $p$ with respective means and covariances $\mu_q, \Sigma_q, \mu_p, \Sigma_p$ can be written as:
\begin{align}
	\label{eqn:KLdiv_multivar}
	&\KL(q||p) = \frac{1}{2}( \underbrace{\log\frac{|\Sigma_p|}{|\Sigma_q|}}_{(a)} -D +\\ 
	&\underbrace{\Tr(\Prec{p}\Sigma_q)}_{(b)} + \underbrace{(\mu_p-\mu_q)^T\Prec{p}(\mu_p-\mu_q)}_{(c)}) \nonumber
\end{align}
The choice of $q$ and $p$ is suggestive. using Eq. \ref{eqn:KLdiv_factorized_app} \& \ref{eqn:KLdiv_multivar}, 
we can derive a closed form for the KL divergence between $q(z_1\ldots z_T)$ and $p(z_1\ldots z_T)$.
$\mu_q,\Sigma_q$ are the outputs of the variational model. Our functional form for $\mu_p,\Sigma_p$ is based on our generative and can
be summarized as: 
\begin{align*}
	&\mu_{p1} = 0\qquad \Sigma_{p1} = \Id\qquad \nonumber\\ 
	&\mu_{pt} = G(z_{t-1},u_{t-1}) = G_{t-1}\qquad \Sigma_{pt} = \dt\vsigma
\end{align*}

Here, $\Sigma_{pt}$ is assumed to be a learned diagonal matrix and $\dt$ a scalar parameter.  

\textbf{Term (a)}
For $t=1$, we have:
\begin{equation}
\label{eqn:logdet_1}
\log\frac{|\Sigma_{p1}|}{|\Sigma_{q1}|} = \log|\Sigma_{p1}|-\log|\Sigma_{q1}| = -\log|\Sigma_{q1}|
\end{equation}

For $t>1$, we have:
\begin{align}
	\label{eqn:logdet_t}
	&\log\frac{|\Sigma_{pt}|}{|\Sigma_{qt}|} = \nonumber\\
	&\log|\Sigma_{pt}|-\log|\Sigma_{qt}|= D \log(\dt) + \log|\vsigma| -\log|\Sigma_{qt}|
\end{align}

\textbf{Term (b)}
For $t=1$, we have:
\begin{equation}
	\label{eqn:trace_1}
	\Tr(\Prec{p1}\Sigma_{q1}) = \Tr(\Sigma_{q1})
\end{equation}

For $t>1$, we have: 
\begin{equation}
	\label{eqn:trace_t}
	\Tr(\Prec{pt}\Sigma_{qt}) = \frac{1}{\dt}\Tr(\diag(\vsigma)^{-1}\Sigma_{qt})
\end{equation}

\textbf{Term (c)}
For $t=1$, we have:
\begin{equation}
	\label{eqn:quad_form_1}
	(\mu_{p1}-\mu_{q1})^T\Sigma_{p1}^{-1}(\mu_{p1}-\mu_{q1}) = ||\mu_{q1}||^2\\
\end{equation}

For $t>1$, we have:
\begin{align}
	\label{eqn:quad_form_t}
	&(\mu_{pt}-\mu_{qt})^T\Sigma_{pt}^{-1}(\mu_{pt}-\mu_{qt}) = \\
	&\dt (G_{t-1}-\mu_{qt})^T\diag(\vsigma)^{-1}(G_{t-1}-\mu_{qt})\nonumber
\end{align}

Rewriting Eq. \ref{eqn:KLdiv_factorized_app} using Eqns. \ref{eqn:logdet_1}, \ref{eqn:logdet_t}, \ref{eqn:trace_1}, \ref{eqn:trace_t}, \ref{eqn:quad_form_1}, \ref{eqn:quad_form_t}, we get:

\begin{align}
	\label{eqn:KLdiv_final}
	&\KL(q(z_1,\ldots,z_T)||p(z_1,\ldots,z_T)) \nonumber \\
	&=\frac{1}{2}((T-1)D \log(\dt)\log|\vsigma| -\sum_{t=1}^T\log|\Sigma_{qt}|  \nonumber\\
	&+ \Tr(\Sigma_{q1})+\frac{1}{\dt}\sum_{t=2}^T \Tr(\diag(\vsigma)^{-1}\Sigma_{qt})+ ||\mu_{q1}||^2 \nonumber\\
	&+ \dt \sum_{t=2}^T \Exp{(G_{t-1}-\mu_{qt})^T\diag(\vsigma)^{-1}(G_{t-1}-\mu_{qt})}{z_{t-1}} )\nonumber
\end{align}

\section{Polyphonic Music Generation}

In the models we trained, the hidden dimension
was set to be $100$ for the emission distribution and $200$ in the transition function.
We typically used RNN sizes from one of $\{400,600\}$ and 
a latent dimension of size $100$. 

\textbf{Samples: } Fig. \ref{fig:poly_samples} depicts mean probabilities of samples
from the \DMM trained on JSB Chorales
\citep{boulanger2012modeling}. 
MP3 songs corresponding to two different samples
from the best \DKF model in the main paper learned on each of the four polyphonic data
sets may be found in the code repository.

\textbf{Experiments with NADE: }
We also experimented with Neural Autoregressive Density Estimators (NADE) \citep{larochelle2011neural} 
in the emission distribution for \DKF-Aug and denote it \DKF-Aug-NADE. 
In Table \ref{tab:polyphonic_nade}, we see that \DKF-Aug-NADE performs comparably to the state of the art RNN-NADE on 
JSB, Nottingham and Piano.  

\begin{table}[h]
	\centering
	\caption{
		\label{tab:polyphonic_nade}
		\small{\textbf{Experiments with NADE Emission: } Test negative log-likelihood (lower is better) on Polyphonic Music Generation dataset.
			\textbf{Table Legend}: RNN-NADE \citep{boulanger2012modeling}}} 
\resizebox{.45\textwidth}{!}{
\begin{tabular}{ccccc}
    \toprule
    ~Methods       & JSB  & Nottingham & Piano & Musedata\\
    \midrule
    \DKF-Aug.-NADE  &  \begin{tabular}{c} 5.118 \\(5.335)\\ \{5.264\}\end{tabular} 
    & \begin{tabular}{c}  2.305 \\(2.347)\\ \{2.364\} \end{tabular}
    & \begin{tabular}{c} 7.048 \\(7.099)\\ \{7.361\} \end{tabular}
    & \begin{tabular}{c} 6.049 \\ (6.115)\\ \{5.247\} \end{tabular}\\
    \midrule
    RNN-NADE & 5.19 & 2.31 & 7.05 & 5.60\\ 
    \bottomrule
    \end{tabular}
    }
\end{table}

\begin{figure}[h]
	\centering
\begin{subfigure}[b]{0.23\textwidth}
	\includegraphics[width=\textwidth,keepaspectratio]{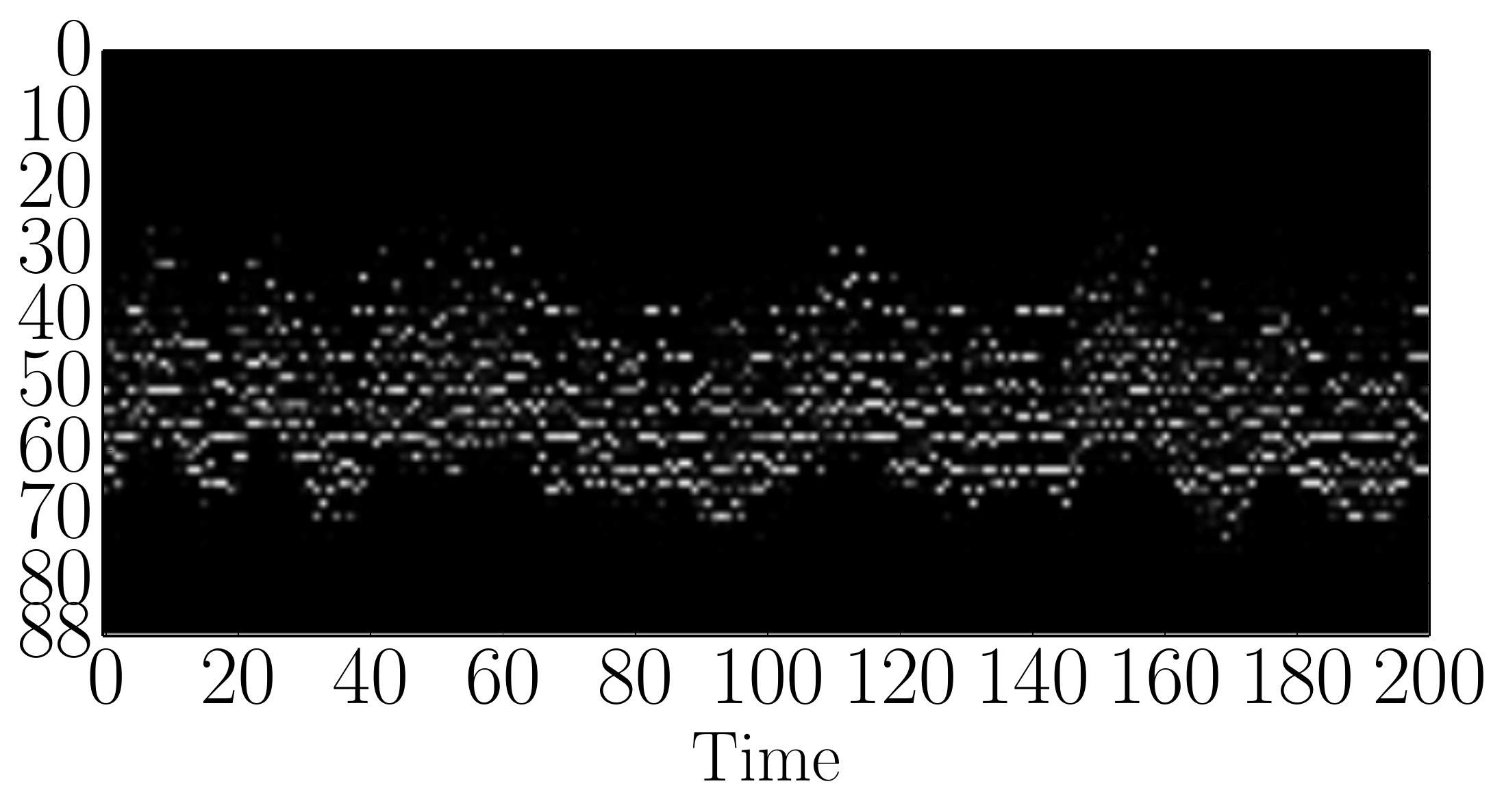}
	\caption{Sample 1}
	\label{fig:jsb-s1}
\end{subfigure}~
\begin{subfigure}[b]{0.23\textwidth}
	\includegraphics[width=\textwidth,keepaspectratio]{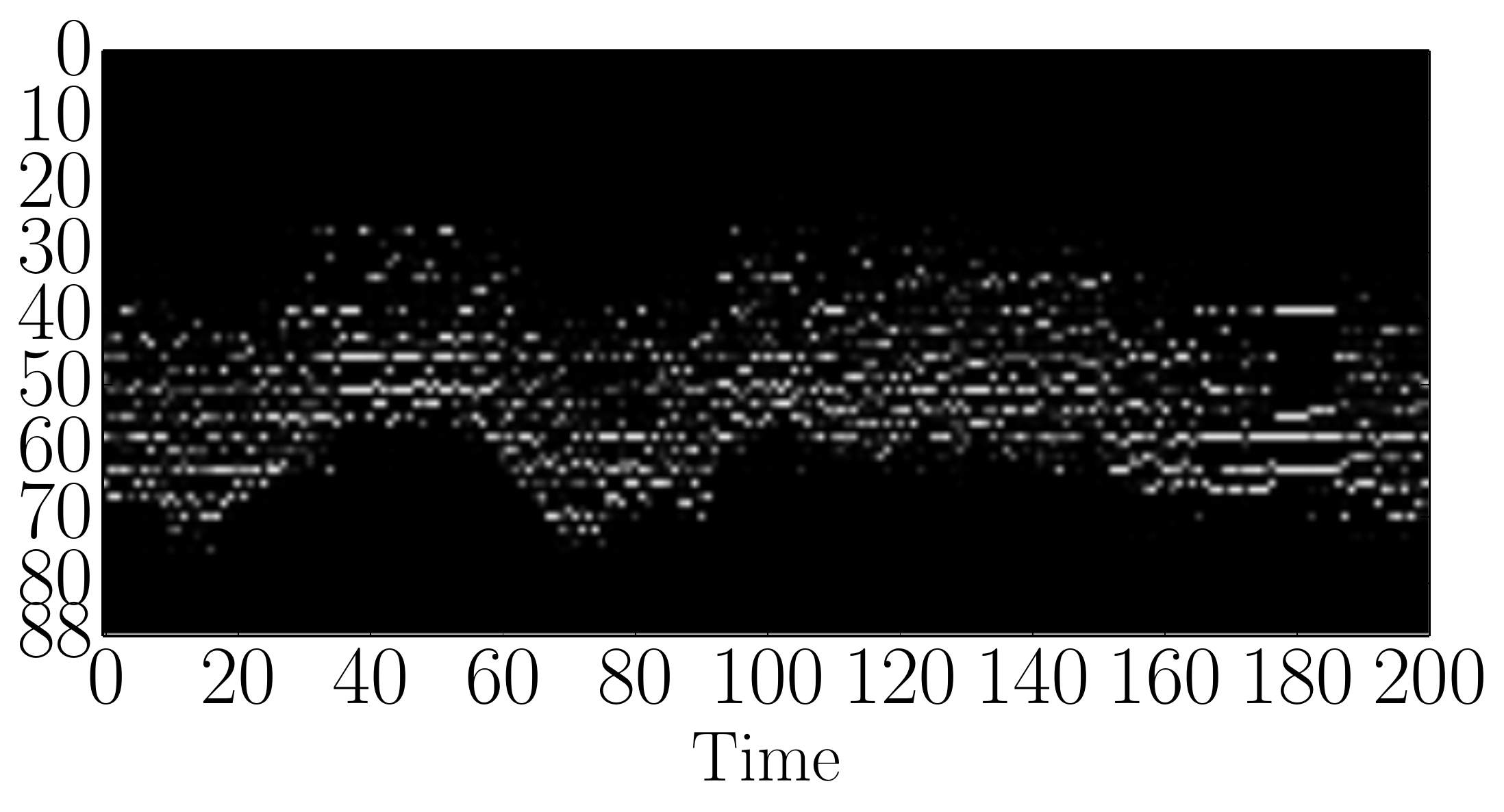}
	\caption{Sample 2}
	\label{fig:jsb-s2}
\end{subfigure}
\caption{Two samples from the \DKF trained on JSB Chorales}
\label{fig:poly_samples}
\end{figure} 

\newpage
\section{Experimental Results on Synthetic Data}

\textbf{Experimental Setup: } We used an RNN size of $40$ in the inference networks used for the synthetic experiments. 

\textbf{Linear SSMs :} Fig. \ref{fig:synthetic9-valid} (N=500, T=25) depicts the performance of inference networks using the same setup as in the main paper, only now using held out data to evaluate the RMSE and the upper bound. 
We find that the results echo those in the training set, and that on unseen data points, 
the inference networks, particularly the structured ones, are capable of generalizing compiled inference. 
\begin{figure}[h]
	\includegraphics[width=0.4\textwidth,keepaspectratio]{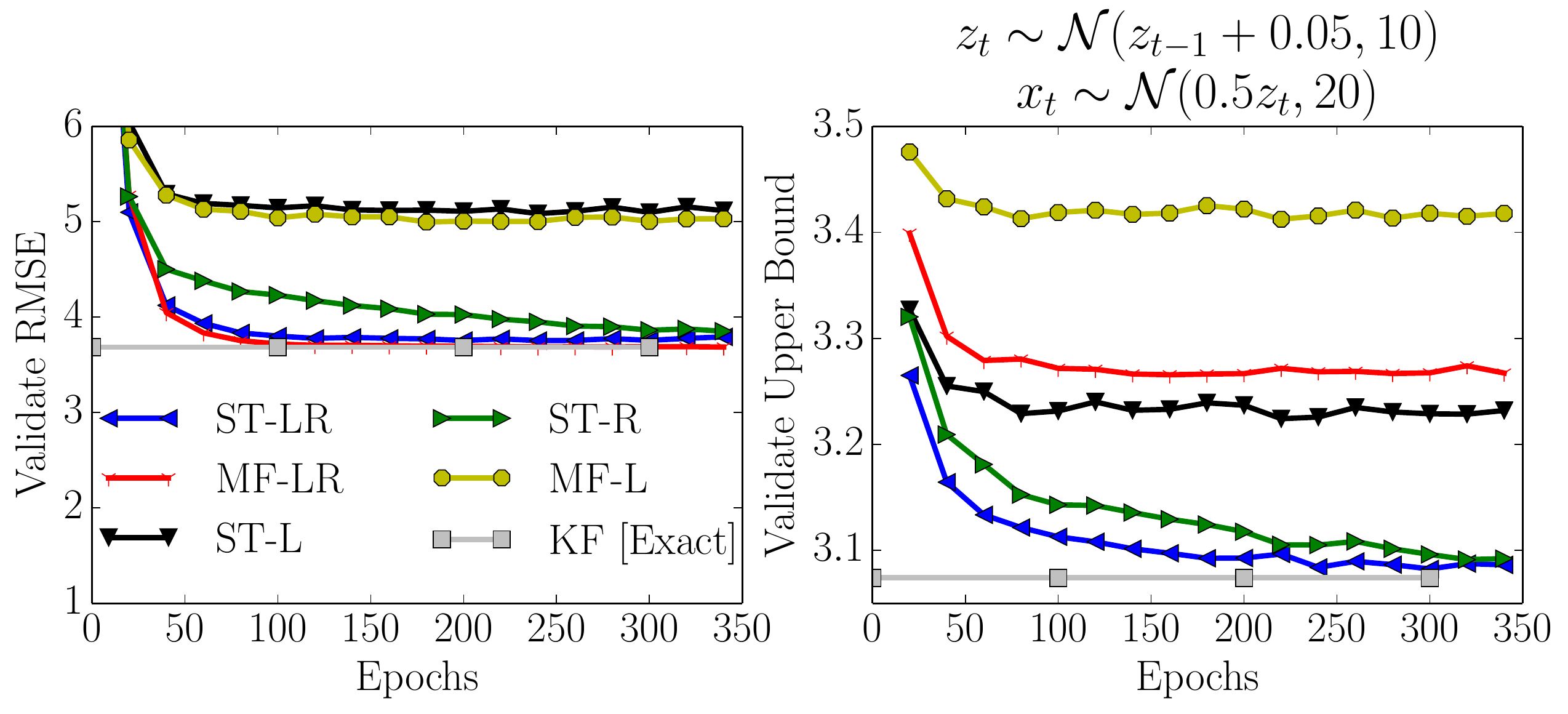}
	\caption{\textbf{Inference in a Linear SSM on Held-out Data: } Performance of inference networks on held-out data using a generative model with Linear Emission and Linear Transition (same setup as main paper)}
	\label{fig:synthetic9-valid}
\end{figure}

\begin{figure}[h]
	\centering
\begin{subfigure}[b]{0.45\textwidth}
	\includegraphics[width=\textwidth,keepaspectratio]{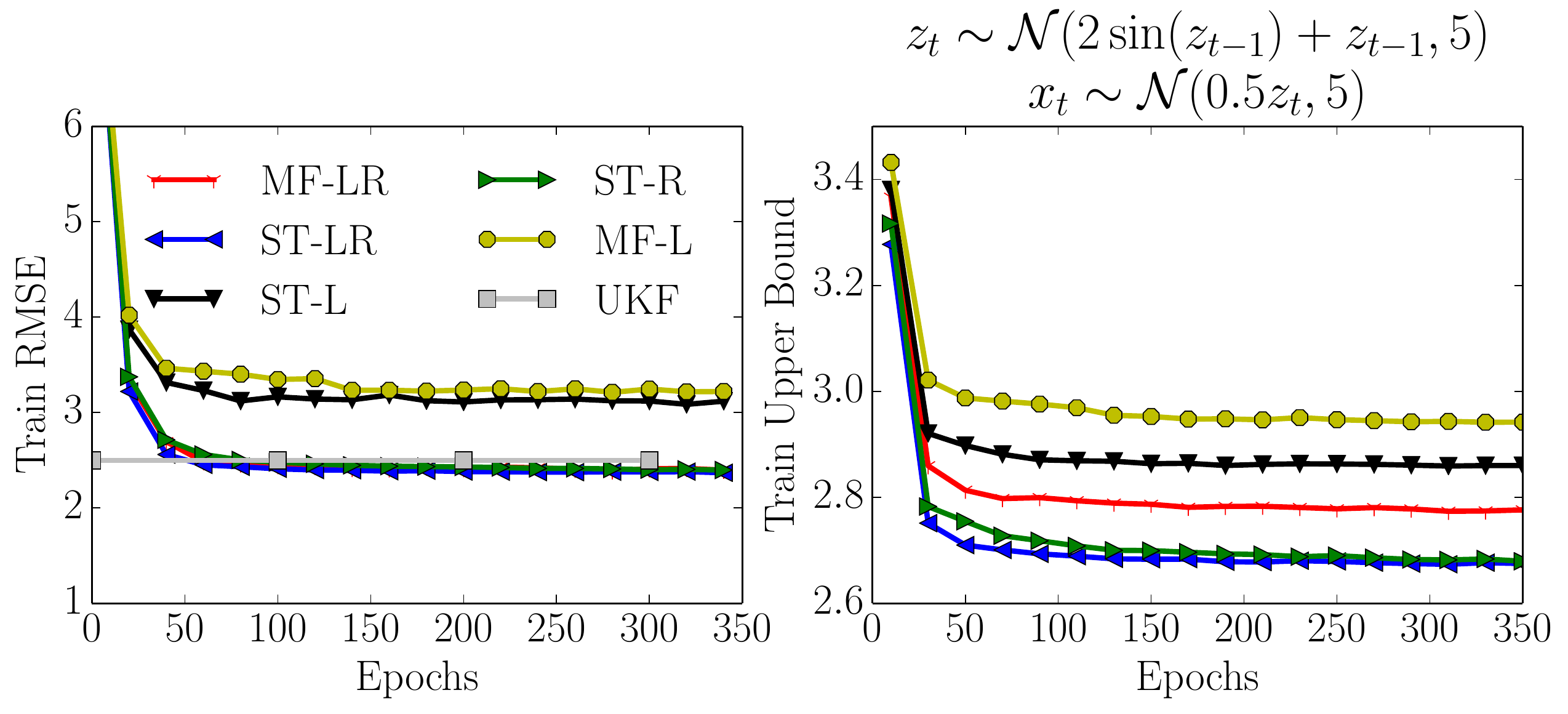}
	\caption{Performance on training data}
	\label{fig:synthetic10-train}
\end{subfigure}
\begin{subfigure}[b]{0.45\textwidth}
	\includegraphics[width=\textwidth,keepaspectratio]{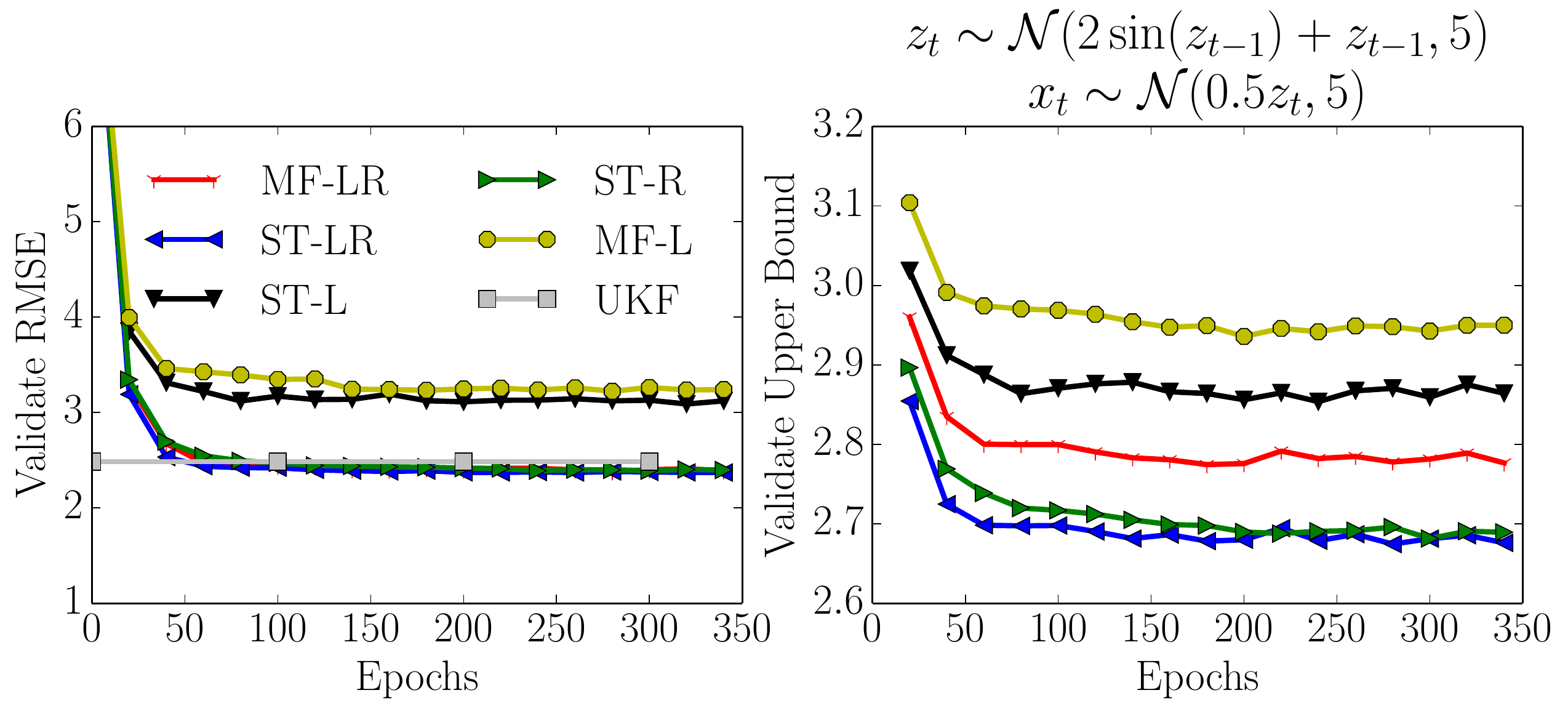}
	\caption{Performance on held-out data}
	\label{fig:synthetic10-valid}
\end{subfigure}
\caption{\textbf{Inference in a Non-linear SSM: } Performance of inference networks trained with data from a Linear Emission and Non-linear Transition SSM}
\label{fig:synth-non-linear}
\end{figure} 

\textbf{Non-linear SSMs :} Fig. \ref{fig:synth-non-linear} considers learning inference networks on a synthetic 
non-linear dynamical system ($N=5000, T= 25$). 
We find once again that inference networks that match the posterior realize
faster convergence and better training (and validation) accuracy. 
 
\begin{figure}[h!]
	\includegraphics[width=0.45\textwidth,keepaspectratio]{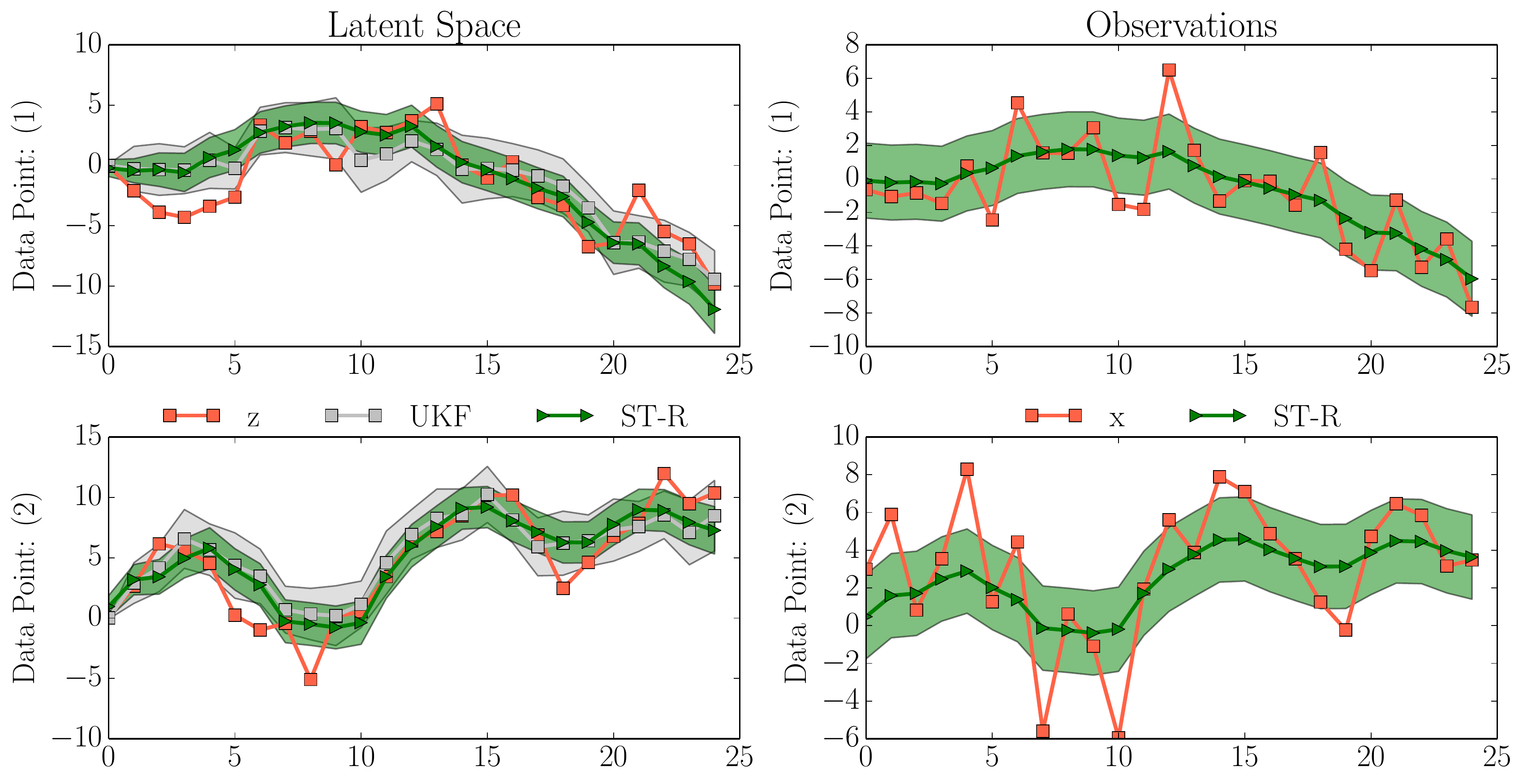}
	\caption{\textbf{Inference on Non-linear Synthetic Data: } Visualizing inference on training data. Generative Models: (a) Linear Emission and Non-linear Transition 
	$z^*$ denotes the latent variable that generated the observation. $x$ denotes the true data. We compare against the results obtained by a smoothed Unscented Kalman Filter (UKF) \citep{wan2000unscented}. 
	The column denoted ``Observations" denotes the result of applying the
emission function of the respective generative model on the posterior estimates shown in the column ``Latent Space".
The shaded areas surrounding each curve $\mu$ denotes $\mu\pm\sigma$ for each plot.
}
\label{fig:synth-reconstructions}
\end{figure} 

\textbf{Visualizing Inference: } In Fig. \ref{fig:synth-reconstructions} 
we visualize the posterior
estimates obtained by the inference network. 
We run posterior inference on the training set $10$ times and take the empirical
expectation of the posterior means and covariances of each method. 
We compare posterior estimates with those obtained by a smoothed 
Unscented Kalman Filter (UKF) \cite{wan2000unscented}. 

\section{Generative Models of Medical Data}

In this section, we detail some implementation details and visualize samples from the generative model trained on patient data. 

\textbf{Marginalizing out Missing Data: } We describe the method we use 
to implement the marginalization operation. The main paper notes that 
marginalizing out observations in the \DMM corresponds to ignoring absent observations during learning. 
We track indicators denoting whether A1C values and Glucose values were observed in the data. These are 
used as markers of missingness. During batch learning, at every time-step $t$, we obtain a matrix $B = \log p(x_t|z_t)$ of size batch-size $\times$ 48, where $48$ is the dimensionality of the observations,
comprising the log-likelihoods of every dimension for patients in the batch. We multiply this with a matrix of $M$. $M$ has the same dimensions as $B$
and has a $1$ if the patient's A1C value was observed and a $0$ otherwise. For dimensions that are never missing, $M$ is always $1$. 

\textbf{Sampling a Patient: }
We visualize samples from the \DMM trained on medical data in Fig. \ref{fig:pat_samples}
The model captures correlations 
within timesteps as well as variations in A1C level and Glucose level across timesteps. 
It also captures rare occurrences of comorbidities found amongst diabetic patients. 

\onecolumn
\begin{figure}[h]
	\includegraphics[width=0.8\linewidth]{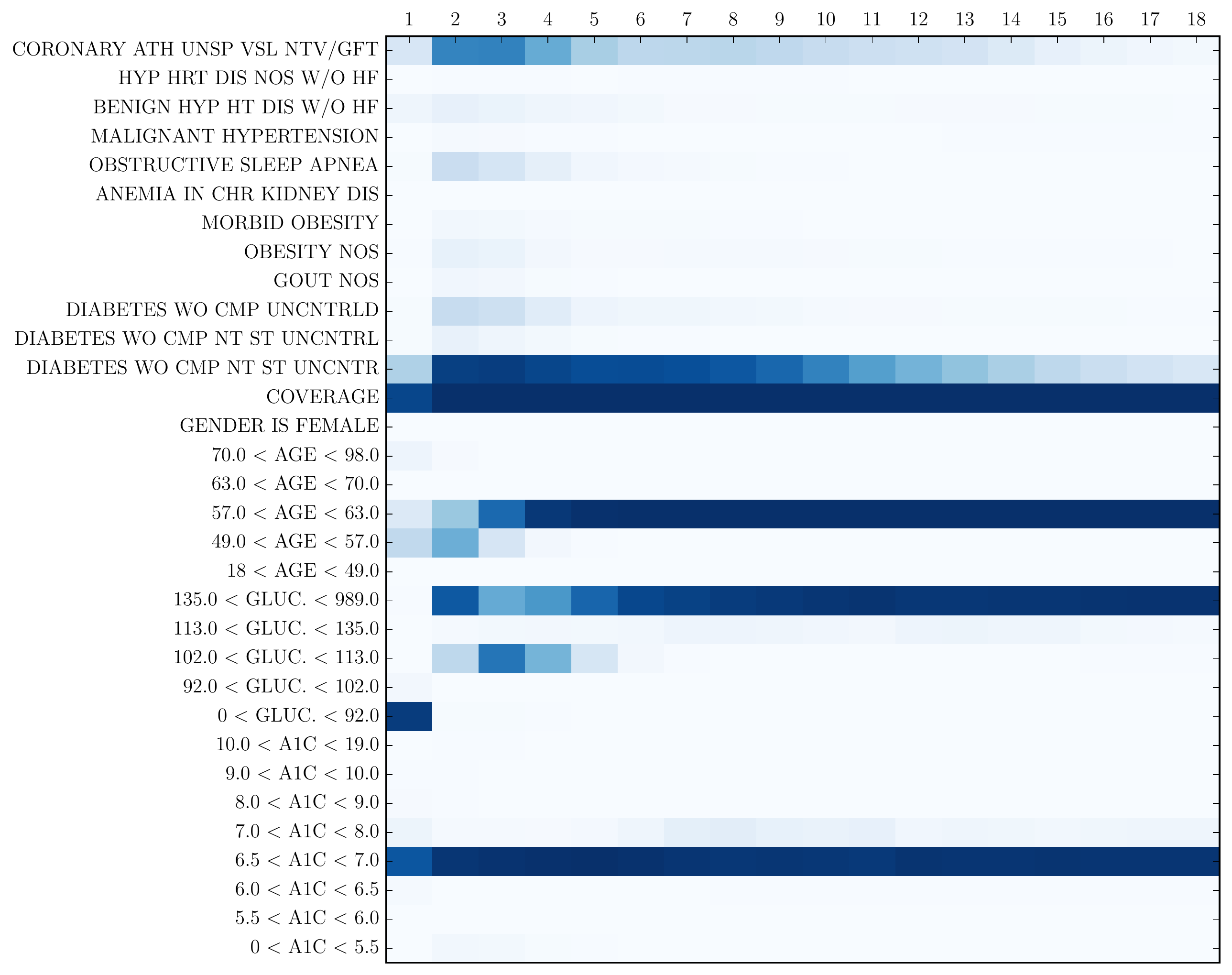}
	\includegraphics[width=0.8\linewidth]{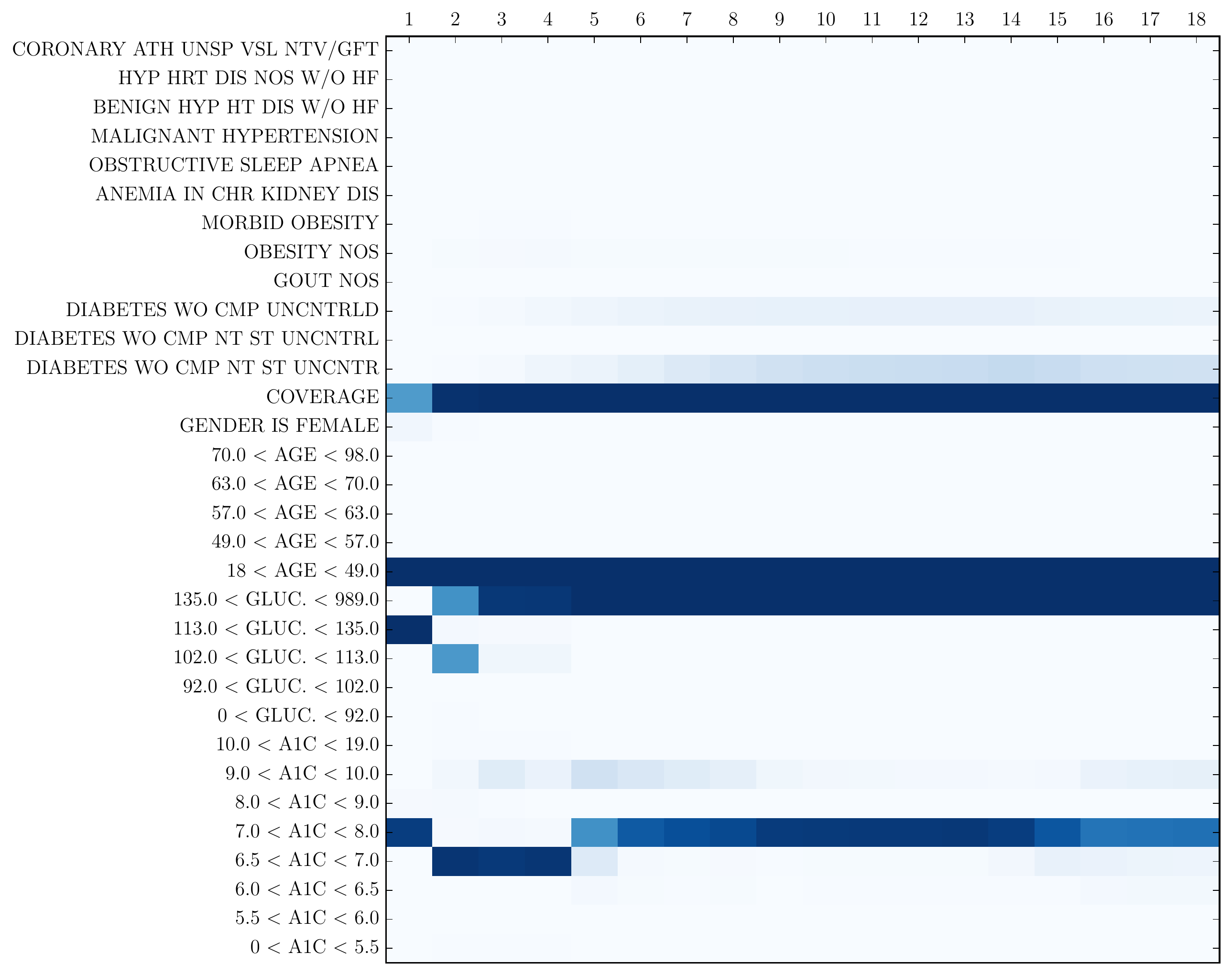}
	\caption{\small
		\textbf{Generated Samples} Samples of a patient from the model, including the most important observations. The x-axis denotes time and the y-axis denotes the observations. The intensity of the color denotes its value between zero and one}
	\label{fig:pat_samples}
\end{figure}
\twocolumn

\end{document}